\documentclass[letterpaper]{article} % DO NOT CHANGE THIS
\usepackage{aaai23}  % DO NOT CHANGE THIS
\usepackage{times}  % DO NOT CHANGE THIS
\usepackage{helvet}  % DO NOT CHANGE THIS
\usepackage{courier}  % DO NOT CHANGE THIS
\usepackage[hyphens]{url}  % DO NOT CHANGE THIS
\usepackage{graphicx} % DO NOT CHANGE THIS
\usepackage{natbib}  % DO NOT CHANGE THIS AND DO NOT ADD ANY OPTIONS TO IT
\usepackage{caption} % DO NOT CHANGE THIS AND DO NOT ADD ANY OPTIONS TO IT
\usepackage{algorithm}
\usepackage{algorithmic}
\usepackage{chngcntr}
\usepackage{newfloat}
\usepackage{listings}
\usepackage[hang,flushmargin]{footmisc}

\DeclareCaptionStyle{ruled}{labelfont=normalfont,labelsep=colon,strut=off} % DO NOT CHANGE THIS
\floatstyle{ruled}
\newfloat{listing}{tb}{lst}{}
\floatname{listing}{Listing}

\usepackage{url}            % simple URL typesetting
\usepackage{amsfonts}       % blackboard math symbols
\usepackage{nicefrac}       % compact symbols for 1/2, etc.
\usepackage{graphicx}
\usepackage{subcaption}
\usepackage{multirow}
\usepackage{booktabs}
\usepackage{xcolor}
\usepackage[implicit=false]{hyperref}

\usepackage{amsmath,amsfonts,bm}

\def\figref#1{figure~\ref{#1}}
\def\Figref#1{Figure~\ref{#1}}

\def\secref#1{sect.~\ref{#1}}
\def\eqref#1{eq.~\ref{#1}}
\def\Eqref#1{Eq.~\ref{#1}}
\def\1{\bm{1}}

\def\rvx{{\mathbf{x}}}
\def\rvy{{\mathbf{y}}}

\def\vtheta{{\bm{\theta}}}

\def\vx{{\bm{x}}}
\def\vy{{\bm{y}}}

\def\mI{{\bm{I}}}

\DeclareMathAlphabet{\mathsfit}{\encodingdefault}{\sfdefault}{m}{sl}
\SetMathAlphabet{\mathsfit}{bold}{\encodingdefault}{\sfdefault}{bx}{n}

\newcommand{\E}{\mathbb{E}}

\newcommand{\R}{\mathbb{R}}

\newcommand{\etal}{\textit{et al}.}

\title{Estimating Regression Predictive Distributions with Sample Networks}
\author {
    Ali Harakeh\textsuperscript{\rm 1, \rm 2,}\footnote{\textbf{ali.harakeh@mila.quebec}} \
    Jordan Hu\textsuperscript{\rm 3, \rm 4} \
    Naiqing Guan\textsuperscript{\rm 4} \
    Steven L. Waslander\textsuperscript{\rm 3, \rm 4} \
    Liam Paull\textsuperscript{\rm 1, \rm 2}
}
\affiliations {
    \textsuperscript{\rm 1}Mila - Quebec AI Institute\\
    \textsuperscript{\rm 2}Université de Montréal\\
    \textsuperscript{\rm 3}University of Toronto Institute for Aerospace Studies\\
    \textsuperscript{\rm 4}University of Toronto\\
}

\begin{document}

\maketitle

\begin{abstract}
Estimating the uncertainty in deep neural network predictions is crucial for many real-world applications. A common approach to model uncertainty is to choose a parametric distribution and fit the data to it using maximum likelihood estimation. The chosen parametric form can be a poor fit to the data-generating distribution, resulting in unreliable uncertainty estimates. In this work, we propose SampleNet, a flexible and scalable architecture for modeling uncertainty that avoids specifying a parametric form on the output distribution. SampleNets do so by defining an empirical distribution using samples that are learned with the Energy Score and regularized with the Sinkhorn Divergence. SampleNets are shown to be able to well-fit a wide range of distributions and to outperform baselines on large-scale real-world regression tasks. \looseness=-1
\end{abstract}

%%%%%%%%%%%%%%%%%%%%%%%%%%%%%%%%%%%%%%%%%%%%%%%%%%%%%%%%%%%%%%%%%
% Intro, Problem Definition, Why we work on this, Contributions %
%%%%%%%%%%%%%%%%%%%%%%%%%%%%%%%%%%%%%%%%%%%%%%%%%%%%%%%%%%%%%%%%%
\section{Introduction}
\label{sec:introduction}
% Small introductory paragraph
Capturing uncertainty in predictions is a crucial ability for machine learning models to be safely used in decision-making systems. Predictive uncertainty is commonly modeled by predicting a probability distribution over the output variable conditioned on the input~\cite{hullermeier2021aleatoric}. In this paper, we tackle the problem of predicting probability distributions for regression tasks using deep neural networks.

% Problem with recent literature 
The most common approach for predicting probability distributions assumes the regression target to follow a particular parametric distribution~\cite{gal2016dropout, lakshminarayanan2017simple, kendall2017uncertainties, skafte2019reliable, liu2019accurate, seitzer2022on}, and uses a neural network to learn its parameters. Such assumptions on the particular form of the predictive distribution limit the modeling capacity of these methods and lead to poor performance when the chosen distribution is a poor fit for the data being modeled. 

\begin{figure}
     \begin{subfigure}{0.23\textwidth}
         \includegraphics[trim={5 5 5 5}, clip,width=\textwidth]{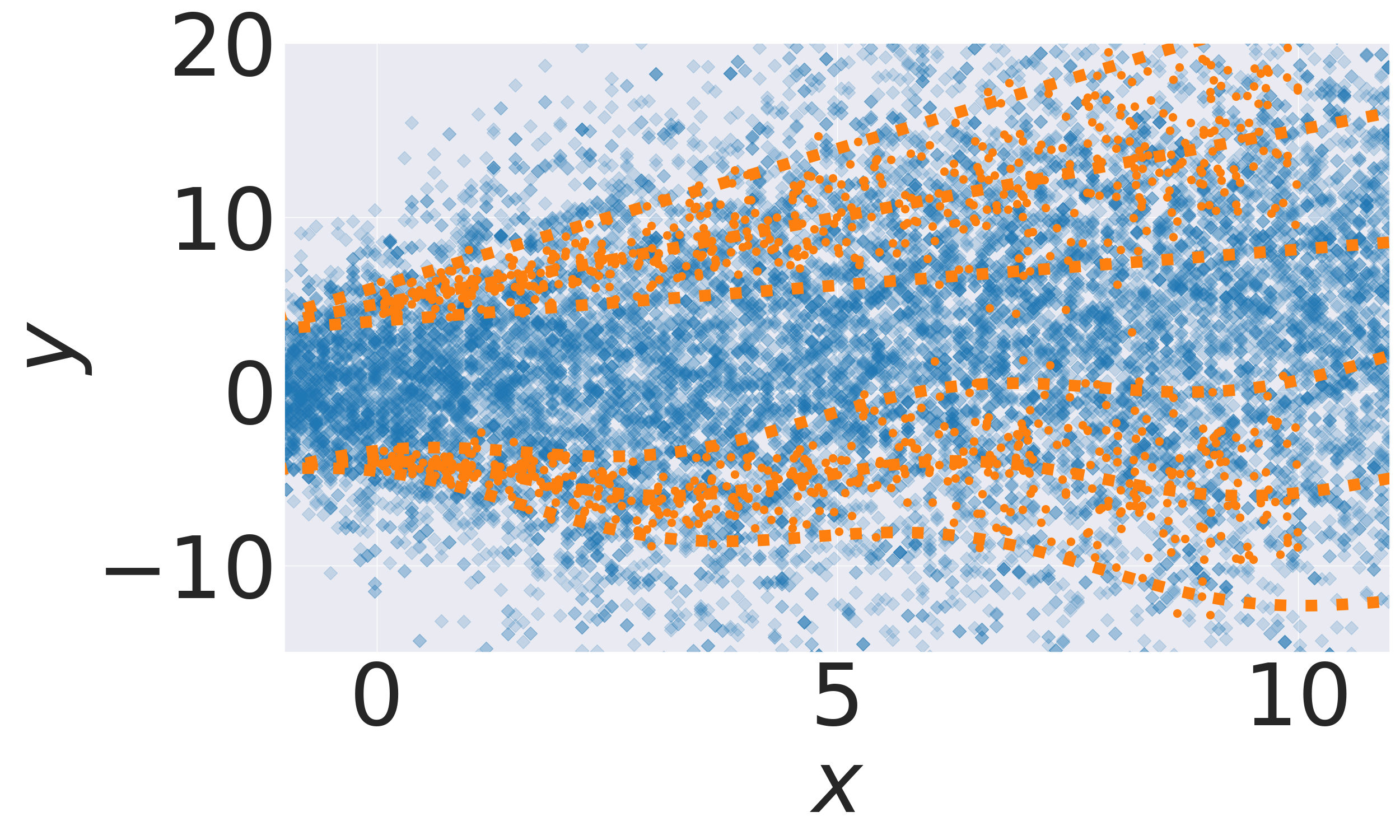}
     \end{subfigure}
    \begin{subfigure}{0.23\textwidth}
         \includegraphics[trim={5 5 5 5}, clip,width=\textwidth]{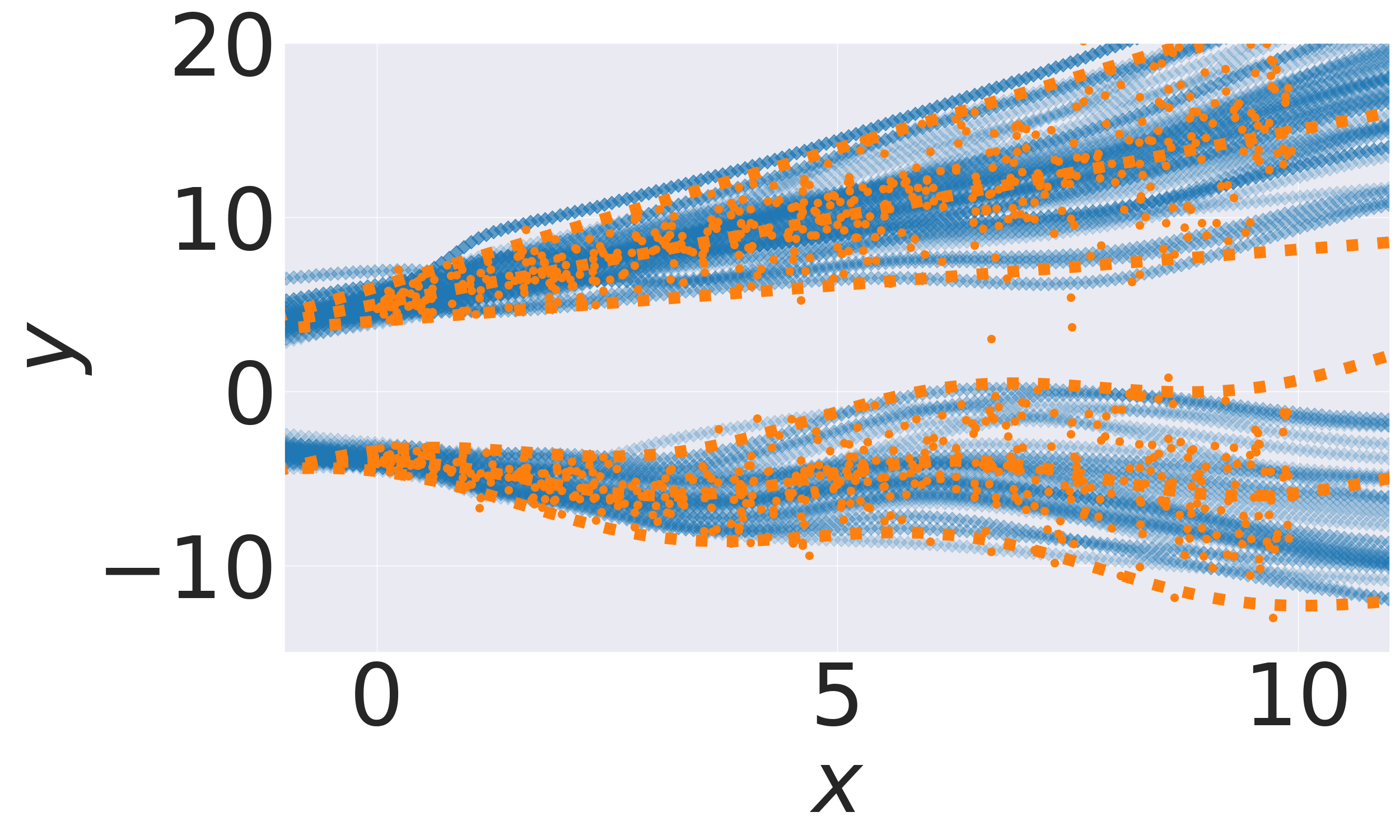}
     \end{subfigure}
      \caption{\textbf{Left:} Samples (\textbf{blue}) from a variance network ($\beta$-NLL)~\cite{seitzer2022on} fails to fit multimodal data (\textbf{orange}). \textbf{Right}: SampleNet produces well-fitted samples.}
     \label{fig:toy_on_sinusoidal_intro}
\end{figure}

This problem is exacerbated by the difficulty in choosing the best parametric distributions for complex, multidimensional real-world data. As an example, using a unimodal Gaussian distribution to model the inherently multimodal movement trajectory of the vehicle at an intersection~\cite{cui2019multimodal} results in a poor approximation of the predictive distribution; we can not expect a neural network to accurately fit multimodal data using an ill-fitting parametric model, even on toy examples (\figref{fig:toy_on_sinusoidal_intro}).

% Our contributions
\textbf{Summary of contributions.} First, we present SampleNet, a simple and scalable neural network architecture that represents a predictive distribution with samples, avoiding the need to choose a particular parametric distribution when modeling predictive uncertainty. Second, we propose to regularize SampleNet by minimizing an optimal transport (OT) cost with respect to a data prior. Our regularization allows us to avoid overfitting while retaining the ability to encourage the predictive distribution to approximate a scaled instance of the prior if needed. Third, we empirically show that SampleNets outperform baselines on real-world multimodal data, standard regression benchmarks, and monocular depth prediction.

%%%%%%%%%%%%%%%%%%%%%%%%%%%%%%%%%%%%%%%%%%%%%%%%%%%%%%%%%%%%
% Literature Review
%%%%%%%%%%%%%%%%%%%%%%%%%%%%%%%%%%%%%%%%%%%%%%%%%%%%%%%%%%%%
\section{Related Work}
\label{sec:related_work}
% Types of Uncertainty and Parameteric Methods
Uncertainty can originate from two inherently different sources~\cite{hullermeier2021aleatoric}, \textit{epistemic uncertainty} which is caused by the lack of knowledge about the best model parameters, and \textit{aleatoric uncertainty} which is caused by the inherent randomness in the underlying process being modeled. Variance networks~\cite{nix1994meanvariance} model aleatoric uncertainty by learning the parameters of a heteroscedastic Gaussian predictive distribution using the negative log-likelihood (NLL) loss. Variance networks have been improved to capture epistemic uncertainty by using them as members of ensembles~\cite{lakshminarayanan2017simple} or by combining them with dropout~\cite{kendall2017uncertainties}. Additionally, higher quality Gaussian predictive distributions have been generated with a Bayesian treatment over the variance~\cite{stirn2020variational}, learning with modified NLL losses~\cite{skafte2019reliable, seitzer2022on}, or fine-tuning with a maximum mean discrepancy loss~\cite{cui2020calibrated}. Other methods replace the Gaussian parametric assumption with a Laplace distribution~\cite{meyer2020learning} or a Student's t-distribution~\cite{skafte2019reliable} to better handle outliers during training.\looseness=-1

% Mixture Density Networks (MDNs), Quantile regression
The variance networks described above limit the output distribution to have a unimodal parametric form. Mixture Density Networks (MDNs)~\cite{bishop1994mixture} relax this assumption by predicting a weighted mixture of Gaussian distributions, at the expense of an added difficulty in optimization due to numerical instability at higher dimensions, overfitting, and mode collapse~\cite{makansi2019overcoming}. Quantile regression~\cite{tagasovska2019single} allows neural networks to model complex distributions by predicting their quantiles. Unfortunately, many applications depend on a representation of the whole distribution, not just a small number quantile ranges.\looseness=-1

% GANs, DiscoNet, and Normalizing flows
Another class of methods avoid specifying a parameteric predictive distribution by generating samples that represent and empirical distribution. Implicit generative models~\cite{mohamed2016learning} and DiscoNets~\cite{bouchacourt2016disco} achieve this goal by accepting noise vectors as additional information to predict multiple samples for every input. These models are hard to train and can suffer from mode collapse~\cite{tagasovska2019single}. Normalizing flows have also been recently employed to estimate aleatoric uncertainty~\cite{si2022autoregressive}, but require carefully designing architectures that represent bijective functions with a tractable Jacobian.\looseness=-1

SampleNets combine the advantages of the methods listed above to estimate aleatoric uncertainty. They avoid setting a parametric form for the predictive distribution by predicting samples from which sample statistics or quantiles can be extracted. They retain the ability to encourage samples to follow a particular parametric form by employing OT regularization, which also helps to avoid overfitting. Finally, SampleNets are easy to implement with few modifications to base architectures.\looseness=-1

\begin{figure*}[t]
     \centering
     \begin{subfigure}[b]{0.24\textwidth}
         \includegraphics[trim={10 10 10 10}, clip,width=\textwidth]{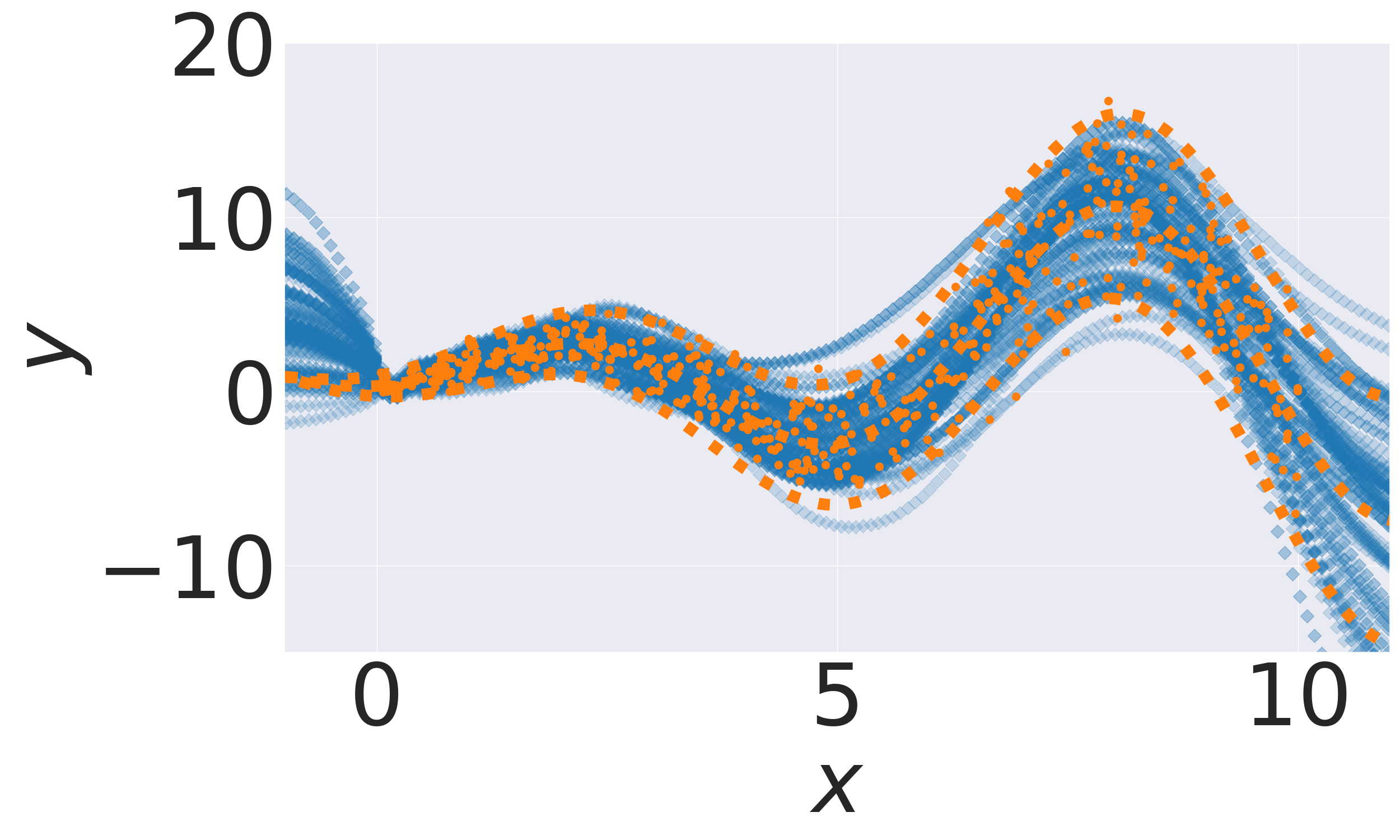}
         \caption{Unimodal Samples}
         \label{fig:unimodal_samples}
     \end{subfigure}
     \begin{subfigure}[b]{0.24\textwidth}
         \includegraphics[trim={10 10 10 10}, clip,width=\textwidth]{png/proper_regularization.png}
         \caption{Multimodal Samples}
         \label{fig:multimodal_samples}
     \end{subfigure}
          \begin{subfigure}[b]{0.24\textwidth}
         \includegraphics[trim={10 10 10 10}, clip,width=\textwidth]{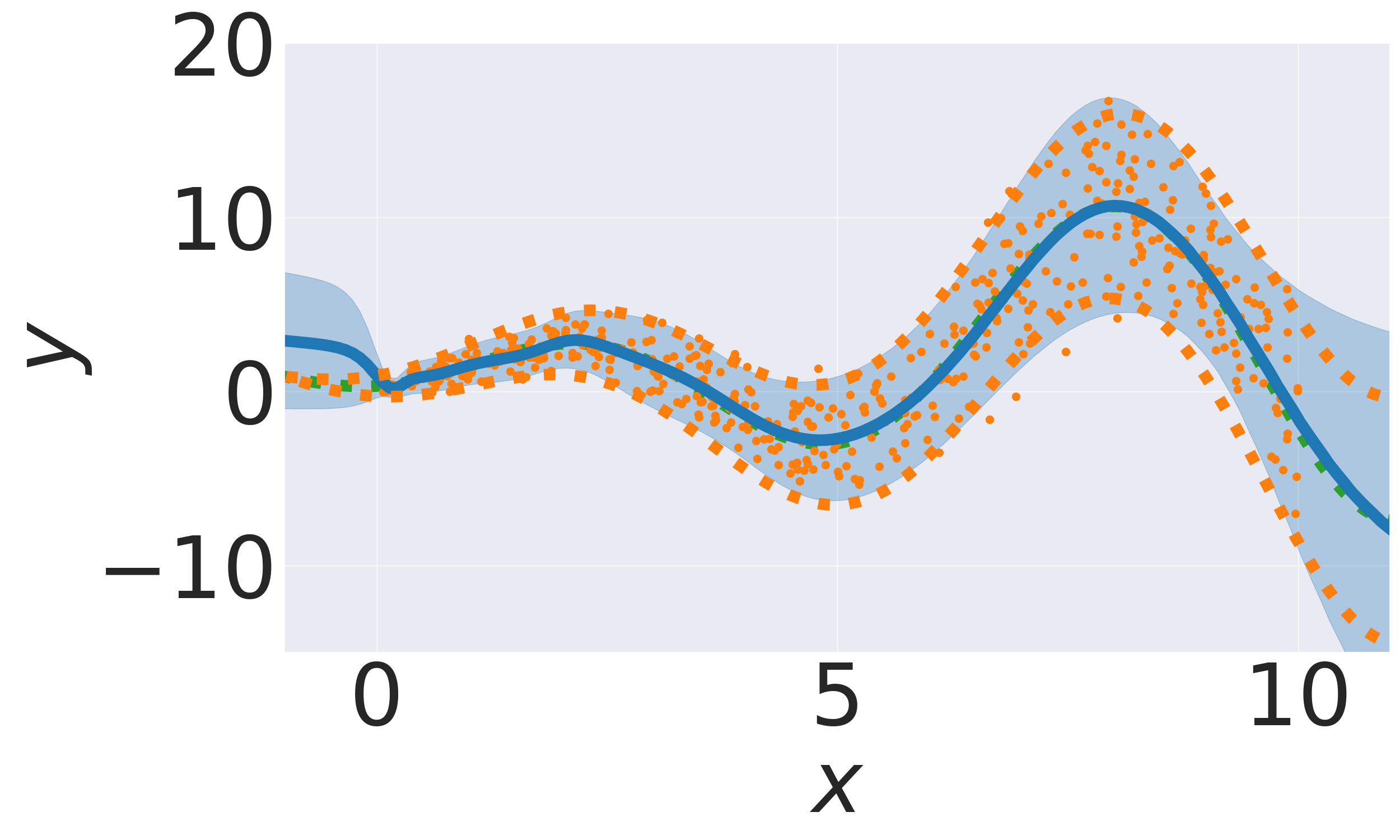}
         \caption{Sample Statistics}
         \label{fig:unimodal_statistics}
     \end{subfigure}
     \begin{subfigure}[b]{0.24\textwidth}
         \includegraphics[trim={10 10 10 10}, clip,width=\textwidth]{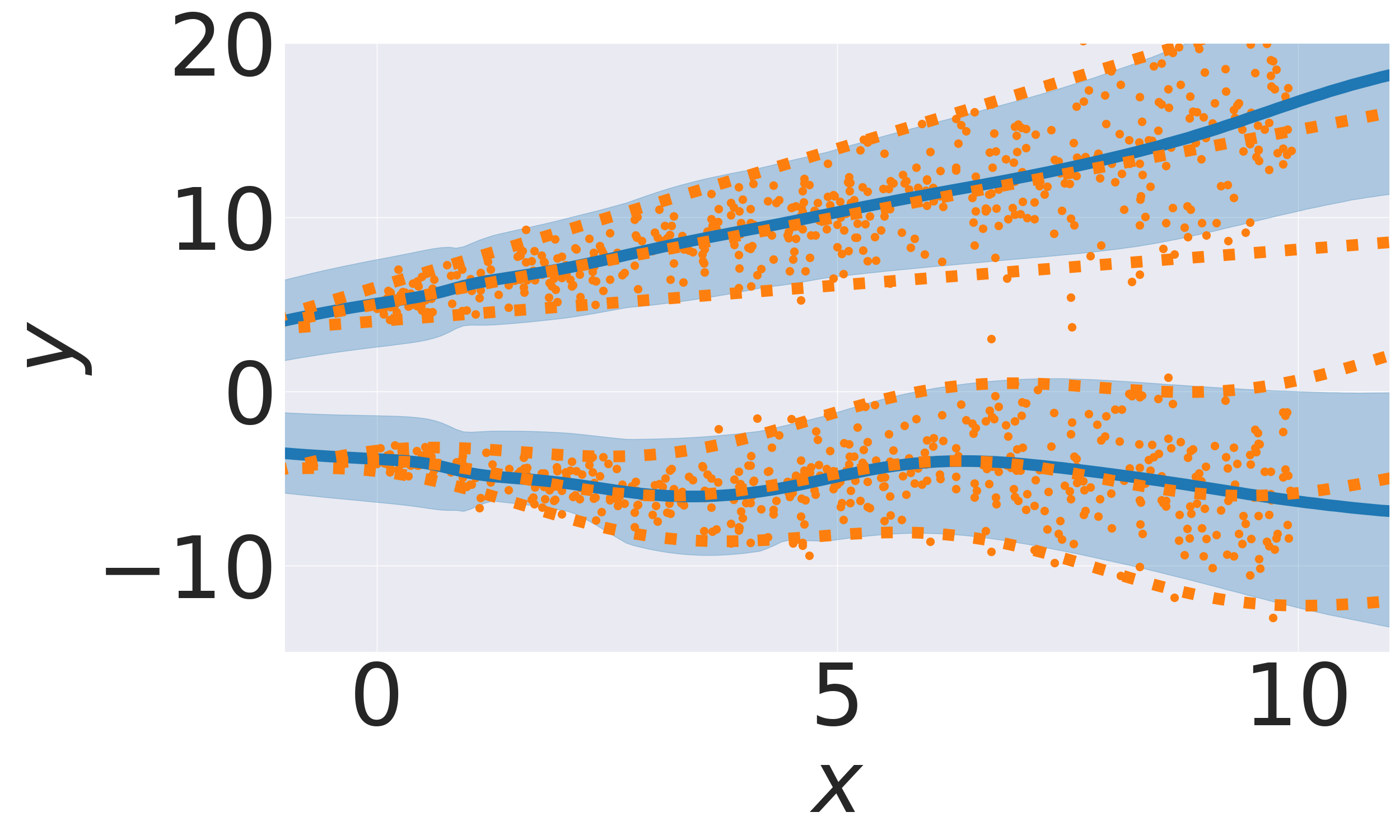}
         \caption{HPD Intervals}
         \label{fig:multimodal_statistics}
     \end{subfigure}
     \caption{\textbf{(a) and (b)}: Samples (\textbf{blue}) predicted by SampleNet to estimate a Gaussian distribution and a multimodal distribution, respectively. Training data and the groundtruth $95\%$ confidence intervals of the data-generating distributions are plotted in \textbf{orange}. \textbf{(c)}: The mean (\textbf{blue line}) and the predicted 95\% confidence interval (\textbf{shaded region}) extracted from samples in (a). \textbf{(d)}: The modes (\textbf{blue line}) and the 75\% highest posterior density (HPD) intervals (\textbf{shaded region}) extracted from samples in (b).
     }
     \label{fig:what_to_do_with_samples}
\end{figure*}

%%%%%%%%%%%%%%%%%%%%%%%%%%%%%%%%%%%%%%%%%%%%%%%%%%%%%%%%%%%%%%
% Problem Definition, Scoring Rules, and Sinkhorn Divergence %
%%%%%%%%%%%%%%%%%%%%%%%%%%%%%%%%%%%%%%%%%%%%%%%%%%%%%%%%%%%%%%
\section{Preliminaries}
\label{sec:preliminaries}
We denote a labelled training dataset of $N$ data pairs as $\mathcal{D} = \{\vx_n, \vy_n\}_{n=1}^{N}$, where $\vx_n \in \R^c$ are input features and $\vy_n \in \R^d$ are output regression targets. The pair $\vx_n,\vy_n$ is assumed to be an i.i.d. sample from the joint data-generating distribution $p^*(\rvy,\rvx)$, and following a true conditional distribution $p^*(\rvy|\rvx)$. Let $p_\vtheta(\rvy|\rvx)$ be a representation of the output predictive (conditional) distribution parameterized by neural network parameters $\vtheta$. Our goal is to learn the parameters $\vtheta$ using $\mathcal{D}$ such that $p_\vtheta(\rvy|\rvx)$ closely estimates $p^*(\rvy|\rvx)$.

\subsection{Proper Scoring Rules}
\label{subsec:proper_scoring_rules}
To learn $p_\vtheta(\rvy|\rvx)$ we minimize a \textit{strictly proper scoring rule} $\mathcal{L}(p_\vtheta(\rvy|\vx_n), \vy_n)$~\cite{lakshminarayanan2017simple}. Under strict properness~\cite{gneiting2007strictly}, the minimum expected score is achieved if and only if $p_\vtheta(\rvy|\rvx) = p^*(\rvy|\rvx)$. Using a proper scoring rule allows us to quantitatively evaluate both the sharpness and the calibration of $p_\vtheta(\rvy|\rvx)$~\cite{gneiting2007probabilistic} on validation and test datasets, where the lower its value, the closer $p_\vtheta(\rvy|\rvx)$ is to $p^*(\rvy|\rvx)$.\looseness=-1

One of the most widely-used strictly proper scoring rules is the negative log-likelihood (NLL), written as
\begin{equation}
\begin{aligned}
         \mathcal{L}_{\text{NLL}}(\vtheta) &= \E\left[-\log p_\vtheta(\rvy|\rvx)\right] \\ &\approx \frac{1}{N}\sum\limits_{n=1}^N -\log p_\vtheta(\vy_n|\vx_n),
\end{aligned}
\label{eq:negative_log_likelihood}
\end{equation}
where $N$ is the number of samples. Efficiently computing the NLL requires specifying a parametric form of the output distribution $p_\vtheta(\vy_n|\vx_n)$. On the other hand, the Energy Score (ES)~\cite{gneiting2007strictly, gneiting2008assessing} is a strictly proper scoring rule written as
\begin{equation}
       \mathcal{L}_{\text{ES}}(\vtheta) = \E || \hat{\rvy} - \rvy|| -  \frac{1}{2}\E || \hat{\rvy} - \hat{\rvy}'||,
\label{eq:energy_score}
\end{equation}
where $||.||$ is the euclidean distance, $\{\hat{\rvy}, \hat{\rvy}'\}\sim p_\vtheta(\rvy|\rvx)$, and $\{\rvy\} \sim p^*(\rvy|\rvx)$. \Eqref{eq:energy_score} shows that the ES is defined using expectations, and as such \textit{does not put restrictions on the parametric form of the predictive distribution to be learned or evaluated}. Given $\hat{\vy}_{n,m} \forall m \in \{1,\dots, M\}$ drawn from $p_\vtheta(\rvy|\vx_n)$ and $\vy_n$ drawn from $p^*(\rvy|\vx_n)$, the ES can be approximated as

\begin{multline}
     \mathcal{L}_{\text{ES}}(\vtheta) \approx \frac{1}{N}\sum\limits_{n=1}^{N} \biggl( \frac{1}{M} \sum\limits_{i=1}^M ||\hat{\vy}_{n,i}-\vy_n|| \\ -\frac{1}{2M^2} \sum\limits_{i=1}^M \sum\limits_{j=1}^M ||\hat{\vy}_{n,i} -\hat{\vy}_{n,j}||\biggr),
      \label{eq:energy_score_samples}
\end{multline}

In \secref{sec:experiments_results} we will be using \eqref{eq:energy_score_samples} to train SampleNet. We will also use \eqref{eq:energy_score_samples} alongside the (Gaussian) NLL in \eqref{eq:negative_log_likelihood} to rank the predictive distributions generated from all methods. For more details on proper scoring rules and the energy score, we refer the reader to~\cite{gneiting2007strictly, harakeh2021estimating}.

\subsection{Geometric Divergences}
\label{subsec:ot_metrics}
Norms and divergences commonly used to compare two probability distributions operate in a point-wise manner, failing to capture the geometric nature of the problem when distributions have non-overlapping supports~\cite{feydy2019interpolating}. Two families of distances that account for the geometry of the underlying space are Maximum Mean Discrepancies (MMD) and Optimal Transport (OT) distances. OT distances have appealing geometric properties~\cite{villani2009optimal}, but suffer from poor sample efficiency while being computationally very expensive. MMDs on the other hand are cheaper to compute and have a small sample complexity, but require significant tuning of their kernel's bandwidth parameter~\cite{genevay2018learning} and can have vanishing gradients~\cite{feydy2019interpolating}.  

Sinkhorn Divergences~\cite{genevay2018learning} are a recently proposed class of geometric divergences that interpolate between OT distances and MMDs~\cite{feydy2019interpolating}, achieving advantages of both by being differentiable, cheap to compute, and scalable to large batch sizes with low sample complexity. We follow~\cite{genevay2018learning} and use the entropy regularized Wasserstein-2 distance, $W_{c}^{\epsilon}(.,.)$~\cite{cuturi2013sinkhorn}, with respect to a ground metric $c$ to define the Sinkhorn Divergence as,
\begin{equation}
S_{c}^{\epsilon}(\boldsymbol{\alpha}, \boldsymbol{\beta})=W_{c}^{\epsilon}(\boldsymbol{\alpha}, \boldsymbol{\beta})-\frac{1}{2}\left(W_{c}^{\epsilon}(\boldsymbol{\alpha}, \boldsymbol{\alpha})+W_{c}^{\epsilon}(\boldsymbol{\beta}, \boldsymbol{\beta})\right),
\label{eq:sinkhorn_divergence}
\end{equation}
where $\boldsymbol{\alpha}, \boldsymbol{\beta}$ are the two probability distributions we want to compare, and $\epsilon>0$ is the entropic regularization strength. In \secref{sec:methodology} we will use \eqref{eq:sinkhorn_divergence} for geometric regularization when training SampleNets, where we use $\epsilon=0.0025$ and $c=2$.\looseness=-1

%%%%%%%%%%%%%%%%%%%%%%%%%
% SampleNet Explanation %
%%%%%%%%%%%%%%%%%%%%%%%%%
\section{SampleNets}
\label{sec:methodology}
The key idea in SampleNets is to predict a set of $M$ samples that represent an empirical predictive distribution $p_\vtheta(\rvy|\rvx)$. This can be achieved by modifying the output layer of any regression architecture to predict $M\times d$ values instead of their usual $1\times d$ output. The samples $\{\hat{\vy}_{n,1} \dots \hat{\vy}_{n,M}\}, \hat{\vy}_{n,i} \in \R^d$ are trained to match the groundtruth sample ${\vy}_{n}$ by minimizing the ES in \eqref{eq:energy_score_samples}.

Predicting samples from $p_\vtheta(\rvy|\rvx)$ eliminates the need for overly restrictive assumptions on the parametric form of $p_\vtheta(\rvy|\rvx)$, resulting in a general architecture that can learn to fit complex (multimodal, skewed) data-generating distributions. We provide a qualitative example in \figref{fig:what_to_do_with_samples}, where the same architecture and hyperparameters are used to predict samples that fit a unimodal Gaussian distribution in \figref{fig:unimodal_samples}, and a multimodal distribution in \figref{fig:multimodal_samples}. A description of the toy data can be found in~\secref{appendix:toy_data}. From predicted samples, we can also estimate summaries of probability distributions such as sample statistics (\figref{fig:unimodal_statistics}) or Highest Posterior Density (HPD)~\cite{hyndman1996computing} intervals (\figref{fig:multimodal_statistics}).\looseness=-1

From \figref{fig:what_to_do_with_samples}, we observe that SampleNets learn $M$ functions of predicted samples over the input space, leading to great flexibility when learning complex distributions. Unfortunately, such flexibility is potentially problematic in that it could lead to a portion of the learned functions overfitting to outliers in training data, particularly when the data-generating distribution is simple. Such overfitting behavior is not specific to SampleNets and can be observed in other over-parameterized architectures like MDNs~\cite{makansi2019overcoming}. A second issue that can arise while training SampleNets is that the sample-based losses used for training suffer from a quadratic memory footprint, causing poor scalability as the number of predicted samples increases. We address these two concerns in sections~\ref{subsec:sinkhorn_regularization} and~\ref{subsec:minibatch_loss_sec} respectively.\looseness=-1
\begin{figure*}[t]
     \centering
     \begin{subfigure}[b]{0.25\textwidth}
         \includegraphics[trim={10 10 10 10}, clip,width=\textwidth]{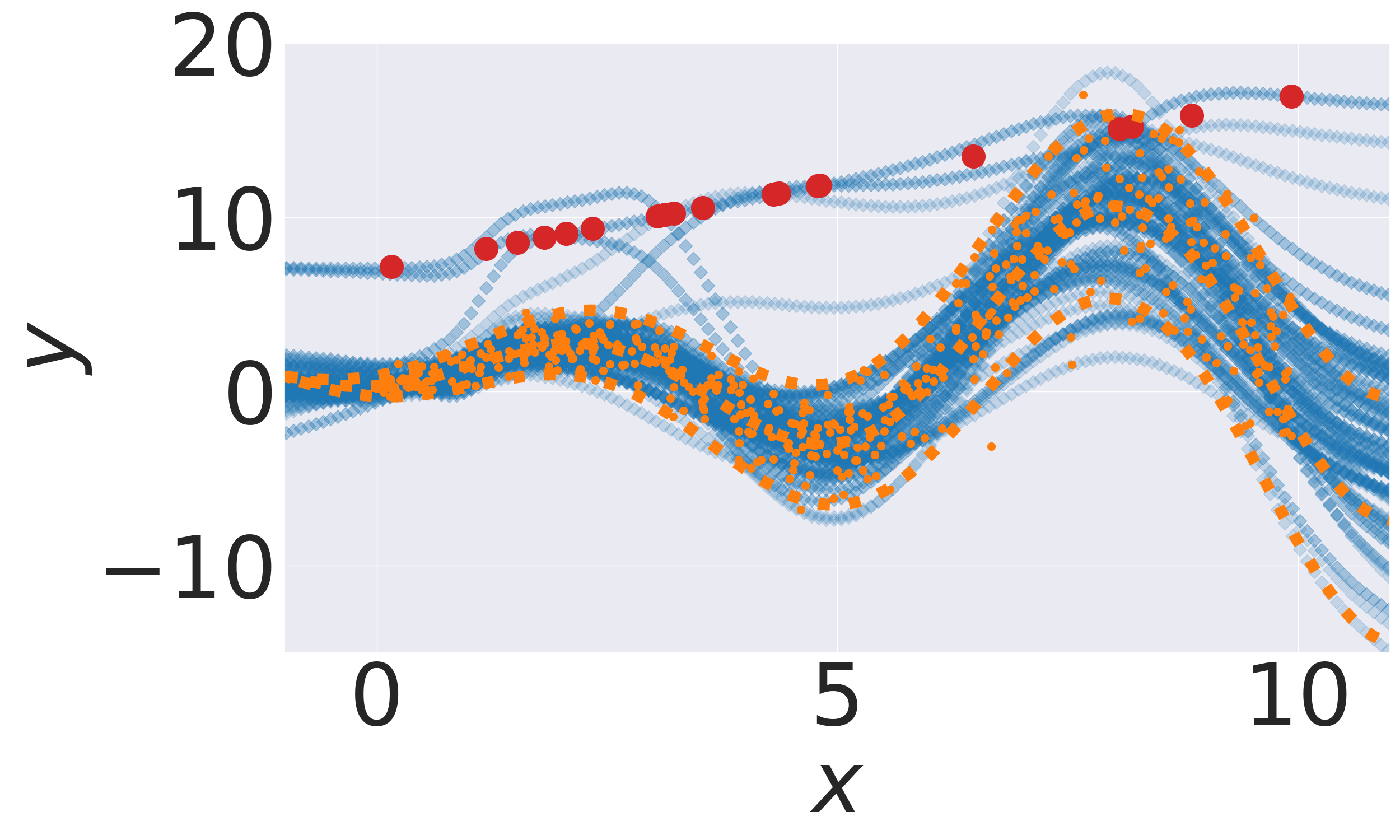}
         \caption{$\eta=0$}
         \label{fig:no_regularization}
     \end{subfigure}
     \begin{subfigure}[b]{0.25\textwidth}
         \includegraphics[trim={10 10 10 10}, clip,width=\textwidth]{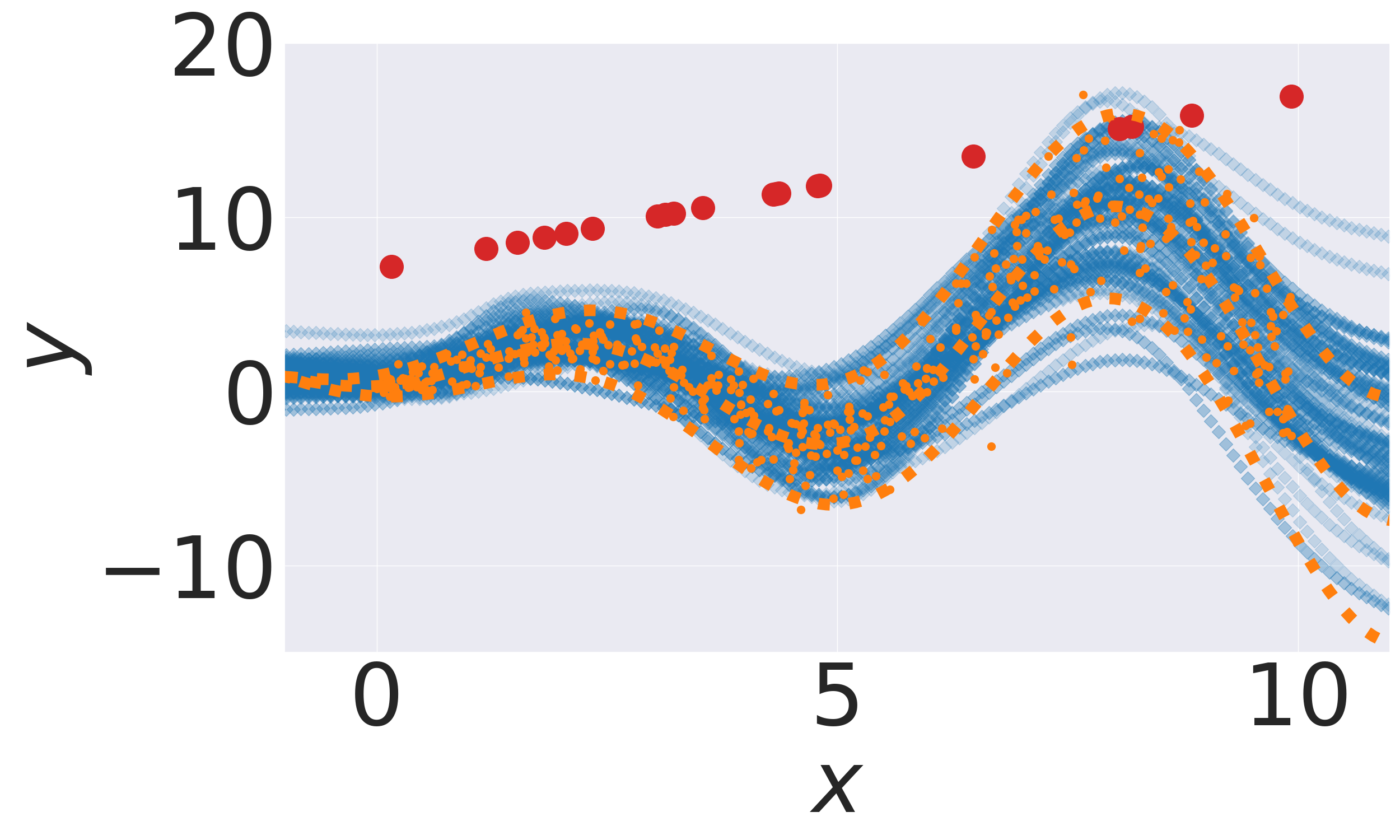}
         \caption{$\eta=2$}
         \label{fig:proper_regularization}
     \end{subfigure}
          \begin{subfigure}[b]{0.25\textwidth}
         \includegraphics[trim={10 10 10 10}, clip,width=\textwidth]{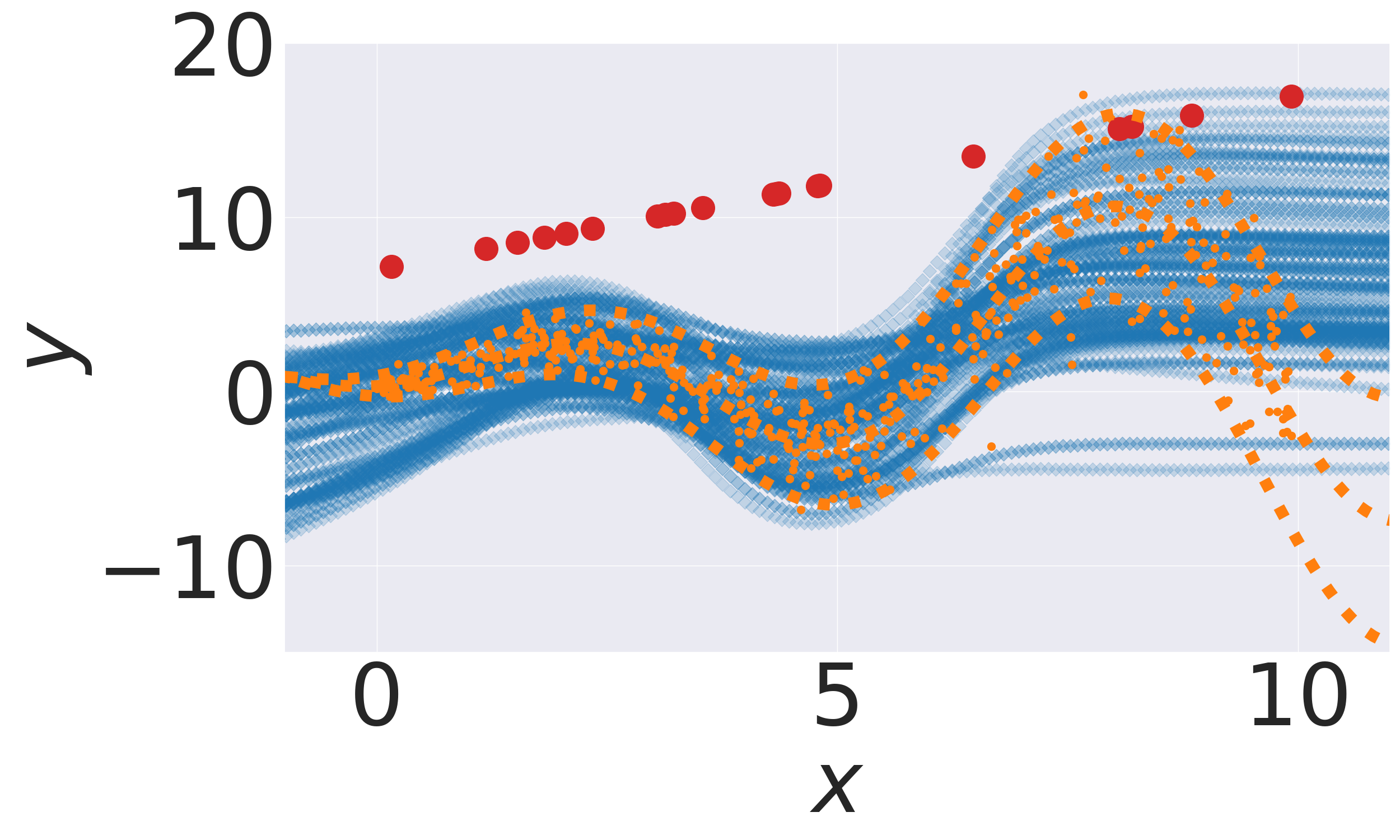}
         \caption{$\eta=200$}
         \label{fig:too_much_regularization}
     \end{subfigure}
     \caption{$M=100$ samples predicted by a SampleNet for various values of $\eta$ in \eqref{eq:total_loss}. The groundtruth distribution and training data are shown in \textbf{orange}. Synthetic outliers are shown in \textbf{red}.
     }
     \label{fig:eta_effect}
\end{figure*}

\subsection{Regularization with the Sinkhorn Divergence}
\label{subsec:sinkhorn_regularization}
The goal of our proposed regularization is twofold. First, we want to avoid the overfitting behavior that originates from overparameterization when training with the ES. Second, we want to preserve the performance that is obtained by state-of-the-art methods when the underlying predictive distribution actually does follow a simple parametric form, e.g. when the data generating distribution is truly Gaussian.\looseness=-1

Inspired by noise contrastive priors~\cite{hafner2020noise}, we propose to constrain the parameters $\vtheta$ of SampleNets to predict samples $\{\hat{\vy}_{n,1}, \dots, \hat{\vy}_{n,M}\}$ by minimizing the optimal transport cost of moving the samples from a data prior $\tilde{\vy}\sim p_{ot}(\rvy|\rvx)$. We achieve this by training SampleNets with a combined loss,
\begin{equation}
\begin{aligned}
     \mathcal{L}_{\text{total}}(\vtheta) &= \mathcal{L}_{\text{ES}}(\vtheta) + \eta \mathcal{L}_{S^\epsilon_c}(\vtheta)\\
      \mathcal{L}_{S^\epsilon_c}(\vtheta) &= \frac{1}{N} \sum\limits_{n=1}^N S^\epsilon_c(p_{ot}(\rvy|\vx_n), p_\vtheta(\rvy|\vx_n)),
      \end{aligned}
     \label{eq:total_loss}
\end{equation}
where $\mathcal{L}_{\text{ES}}$ is the ES loss from \eqref{eq:energy_score_samples}, $S^\epsilon_c$ is the Sinkhorn Divergence from \eqref{eq:sinkhorn_divergence}, and $\eta$ is a hyperparameter that sets the strength of the regularization term. 

We can choose $p_{ot}(\rvy|\vx)$ to be a standard distribution (Uniform, Gaussian) and then normalize the predicted samples $\{\hat{\vy}_{n,1}, \dots \hat{\vy}_{n,M}\}$ before computing the Sinkhorn Divergence. Normalization allows us to decouple the choice of the prior from the magnitude of the specific problem at hand. Specifically, normalization allows the Sinkhorn Divergence in~\ref{eq:total_loss} to be minimized when used with a standard prior distribution as long as the predicted samples follow a scaled version of this prior. The Sinkhorn Divergence only penalizes the shape of the output distribution as it diverges from the prior, preventing the regularization loss from overly restricting the support of the output distribution. More details on normalization can be found in \secref{appendix:samplenet}.\looseness=-1

We visualize the impact of our proposed regularization in \figref{fig:eta_effect} when using a standard Gaussian prior distribution. Using the same choice of hyperparameters, we train a SampleNet to predict $M=100$ samples using $500$ training data points from a unimodal Gaussian distribution (orange). Additionally, we append 20 synthesized outliers to the training data, which we show in red. \Figref{fig:no_regularization} shows that when training SampleNet with $\eta=0$, some of the $100$ learned functions are assigned to fit the outliers in training data. When regularization with ($\eta=2$) (\figref{fig:proper_regularization}) is added, the overfitting behavior is reduced as the optimal transport cost penalizes the predicted samples from fitting the training data outliers, while simultaneously pushing those to follow a Gaussian distribution. On the other hand, \figref{fig:too_much_regularization} shows that at a large value of $\eta=200$, the regularization loss overwhelms the total loss in~\eqref{eq:total_loss}, which leads to the output samples underfitting the training data. Specifically, at a high regularization strength, the output samples are pushed to ignore the geometry of the training data as the total loss can sufficiently be minimized by predicting an arbitrary Gaussian distribution for any input $\vx_n$. We treat $\eta$ as a tunable hyperparameter and study its impact in \secref{sec:experiments_results}.

\subsection{Reducing Compute Costs with Minibatch Losses}
\label{subsec:minibatch_loss_sec}
The sample-based losses in equations~\ref{eq:energy_score_samples} and~\ref{eq:sinkhorn_divergence} suffer from a quadratic compute time and memory growth as a function of $M$. To resolve this issue, we propose to use minibatch subsampling proposed by Genevay \etal~\cite{genevay2018learning} to generate unbiased estimators for the ES and the Sinkhorn Divergence that are computed with a lower sample size $K < M$. Given a number of repetitions $L > 0$, we write the minibatch ES loss as,
\begin{multline}
     \mathcal{L}_{\text{ES}}(\theta) = \frac{1}{N L}\sum\limits_{n=1}^{N}\sum\limits_{l=1}^{L} \biggl( \frac{1}{K} \sum\limits_{i=1}^K ||\hat{\vy}^{(l)}_{n,i}-\vy_n|| \\ -\frac{1}{2K^2} \sum\limits_{i=1}^K \sum\limits_{j=1}^K ||\hat{\vy}^{(l)}_{n,i} -\hat{\vy}^{(l)}_{n,j}||\biggr),
    \label{eq:minibatch_energy_score}   
\end{multline}
where the samples $\hat{\vy}^{(l)}_{n,i} \forall i \in \{1, \dots, K \}$ are the $K$ elements sampled without replacement from $\{\hat{\vy}_{n,1}, \dots \hat{\vy}_{n,M}\}$ during repetition $l$. Similarly, we can write the minibatch subsampled Sinkhorn Divergence loss as:
\begin{multline}
    \mathcal{L}_{S^\epsilon_c}(\vtheta) = \frac{1}{N L}\sum\limits_{n=1}^{N}\sum\limits_{l=1}^{L} S^\epsilon_c(\{\tilde{\vy}^{(l)}_{n,1}, \dots \tilde{\vy}^{(l)}_{n,K}\}, \\ \{\hat{\vy}^{(l)}_{n,1}, \dots \hat{\vy}^{(l)}_{n,K}\}).
    \label{eq:minibatch_sinkhorn}
\end{multline}
The ES originates from the Energy Distance~\cite{rizzo2016energy}, which in turn is an MMD~\cite{sejdinovic2013equivalence} and as such has unbiased gradient estimates when used with minibatch subsampling~\cite{genevay2018learning}. Fatras \etal~\cite{fatras2019learning} studied the theoretical properties of the minibatch Sinkhorn Divergence, which was also shown to have unbiased gradients with respect to model parameters. We empirically study the impact of the values of $M$, $K$, and $L$ on the performance of SampleNets in \secref{sec:experiments_results}.

%%%%%%%%%%%%%%%%%%%%%%%%%%%
% Experiments and Results %
%%%%%%%%%%%%%%%%%%%%%%%%%%%
\section{Experiments and Results}
\label{sec:experiments_results}
Details of our experimental setup can be found in \secref{appendix:experimental_setup}. We perform experiments to answer three questions and draw the following conclusions.

\textbf{Can SampleNet estimate multimodal probability distributions?} We test SampleNet on two real-world datasets with data generating distributions that are multimodal and show its ability to accurately predict these distributions with no assumptions on their parametric form.

\textbf{How does SampleNet perform in comparison to distributional regression baselines?
} We compare SampleNet to other lightweight distributional regression methods on real-world regression datasets and on monocular depth prediction. SampleNet is shown to perform on par or better than all tested baselines.

\textbf{How sensitive is SampleNet to its hyperparameters?}
We study the impact of the number of samples $M$, the minibatch sample size $K$, the number of repetitions $L$, and the regularization strength $\eta$ on the performance of SampleNets when predicting multimodal distributions. We observe the performance of SampleNet to be most sensitive to the regularization strength $\eta$, which we recommend being prioritized during hyperparameter optimization.  We also observe that SampleNet produces high-quality predictive distributions with a small number of samples $M$, increasing $M$, $L$, and $K$ does not lead to a substantial increase in performance on most of our real-world datasets.

\begin{table}[b]
   \centering
   \caption{Comparison with baselines on multimodal datasets. We report the mean ES (lower is better) $\pm$ the standard deviation over $5$ random seeds.}
    \resizebox{0.45\textwidth}{!}{
   \begin{tabular}{lccc}
    \label{table:toy_final_result}
    Dataset &$\beta$-NLL & MDN & SampleNet \\ \cmidrule(lr){1-1} \cmidrule(lr){2-4}
    Weather& 0.594 $\pm$ 0.032& 0.210 $\pm$ 0.065& \textbf{0.106} $\pm$\textbf{0.002} \\ \midrule
    Traffic &133.884 $\pm$ 3.138& 117.208 $\pm$ 4.169& \textbf{12.562} $\pm$ \textbf{0.389}\\
    \bottomrule
   \end{tabular}}
\end{table}

\begin{figure*}[t]
    \centering
     \begin{subfigure}[b]{0.23\textwidth}
         \includegraphics[trim={5 5 5 5}, clip,width=\textwidth]{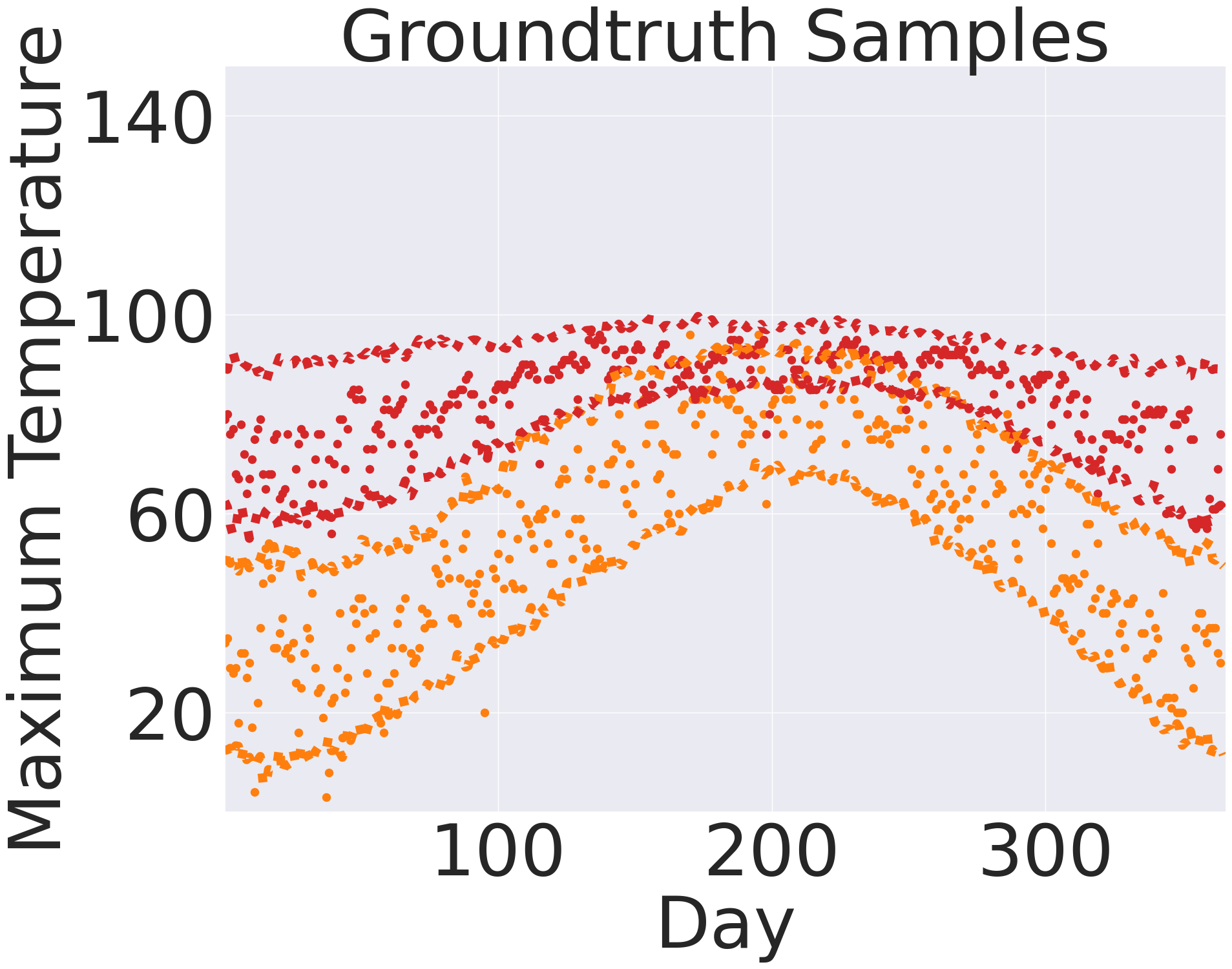}
     \end{subfigure}
          \begin{subfigure}[b]{0.23\textwidth}
         \includegraphics[trim={5 5 5 5}, clip,width=\textwidth]{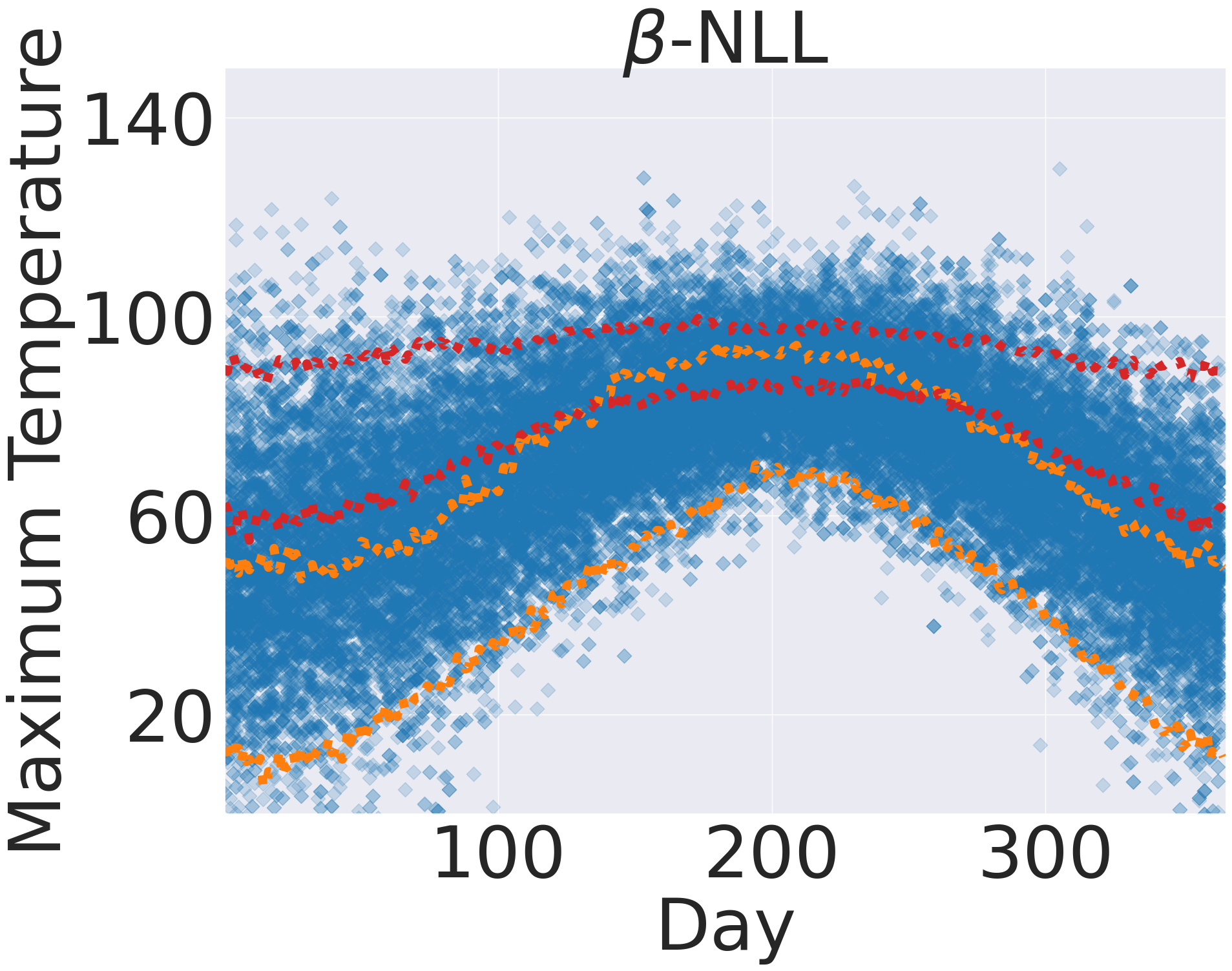}
     \end{subfigure}
          \begin{subfigure}[b]{0.23\textwidth}
         \includegraphics[trim={5 5 5 5}, clip,width=\textwidth]{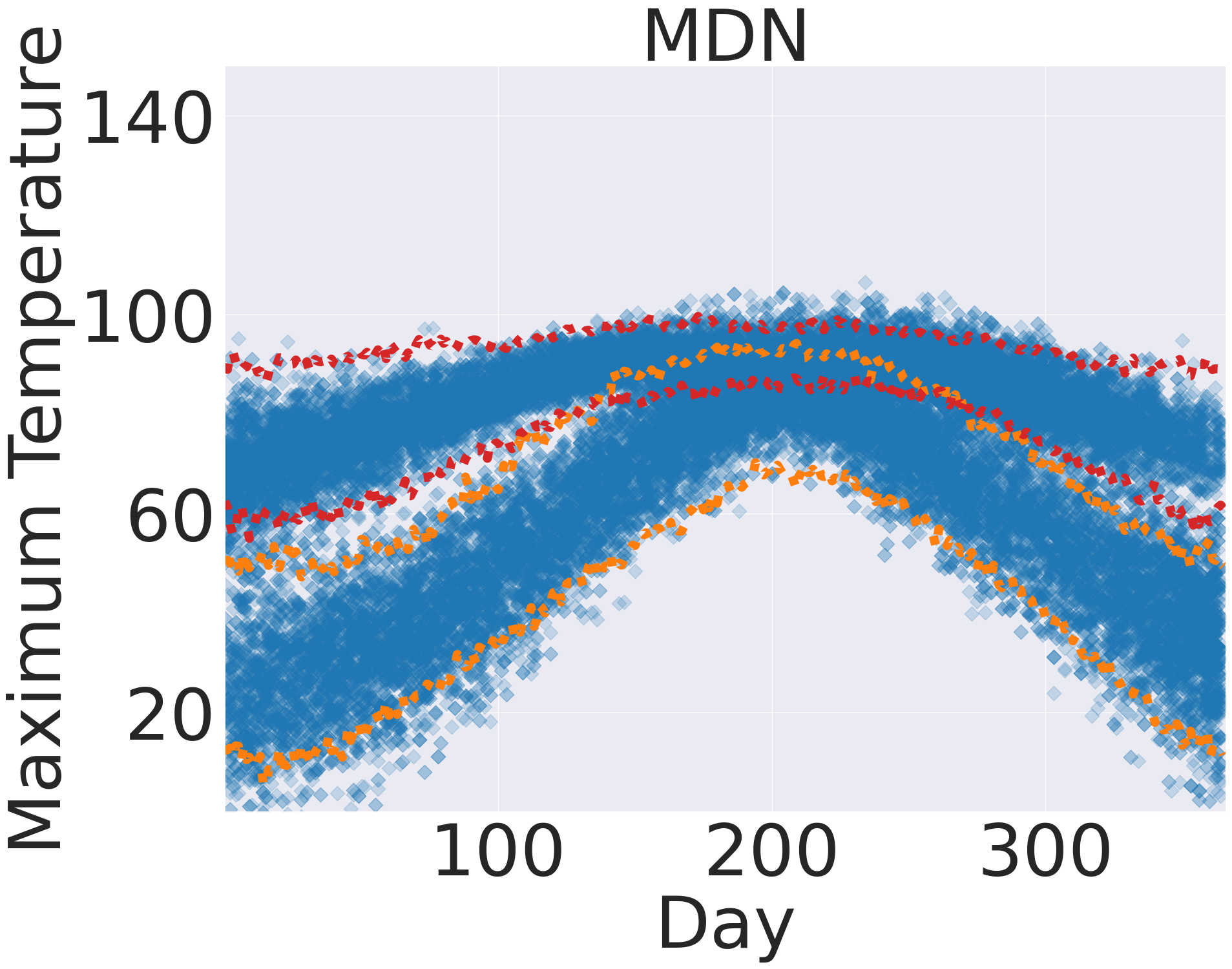}
     \end{subfigure}
          \begin{subfigure}[b]{0.23\textwidth}
         \includegraphics[trim={5 5 5 5}, clip,width=\textwidth]{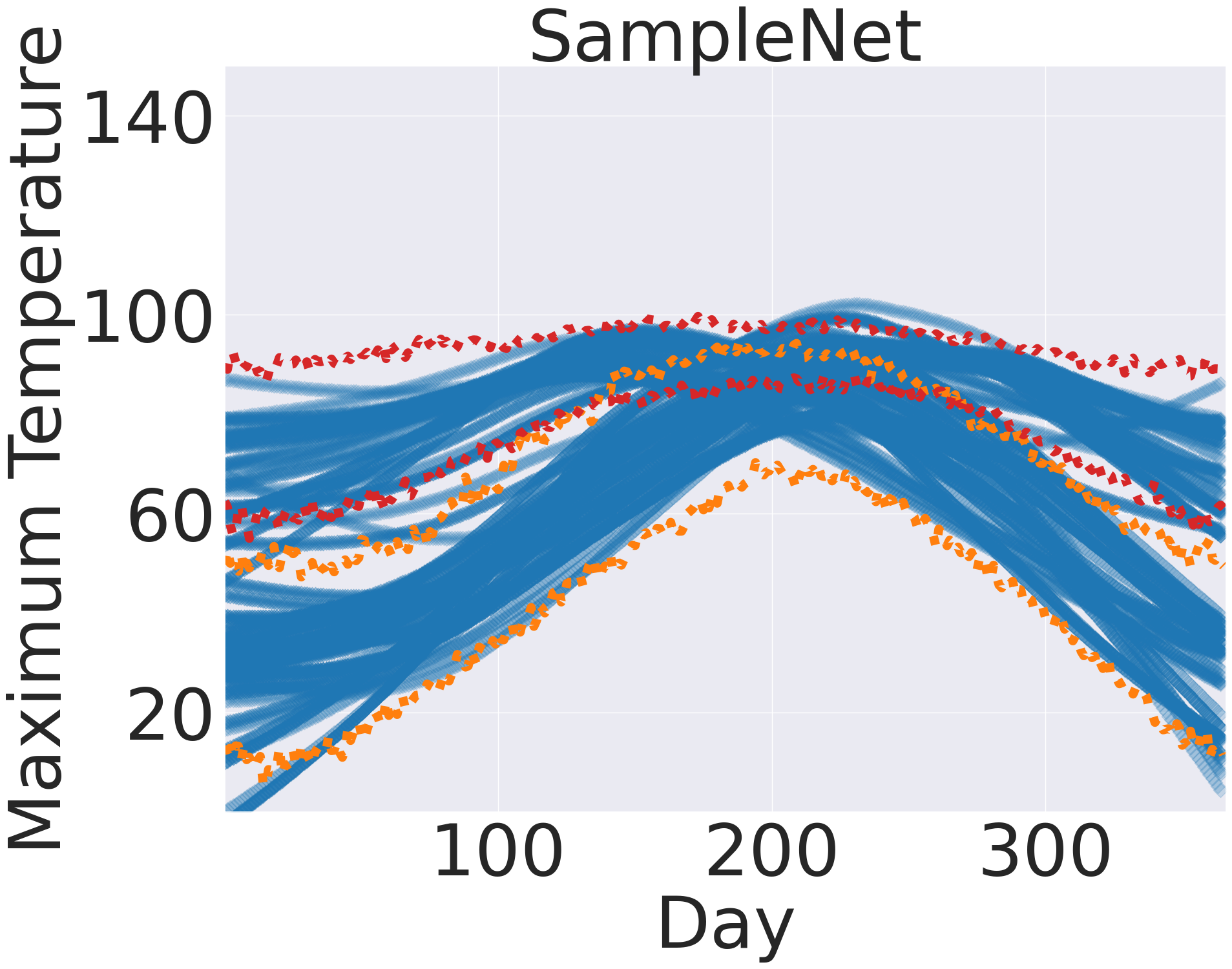}
     \end{subfigure}
      \caption{Scatter plots of samples (\textbf{blue}) from SampleNet and baselines on the Weather dataset. The test set 95$\%$ confidence intervals of maximum daily temperatures from two different weather stations is shown in \textbf{red} and \textbf{orange}. Groundtruth samples used for training are shown in the leftmost plot.}
     \label{fig:weather_qualitative}
\end{figure*}

\subsection{Multimodal Datasets}
To determine SampleNet's capacity to learn complex probability distributions, we perform experiments on the Weather (\figref{fig:weather_qualitative}) and Traffic datasets (\figref{fig:traffic_qualitative}), both of which have been  demonstrated to be multimodal. More details on both datasets as well as the training hyperparameters can be found in appendix~\ref{appendix:multimodal_datasets}. Since the datasets are not Gaussian, we use the ES~(\eqref{eq:energy_score_samples}) as an evaluation metric. We repeat all experiments using $5$ random seeds and report the standard deviation.\looseness=-1 

Table~\ref{table:toy_final_result} shows a quantitative comparison of SampleNet, MDN, and $\beta$-NLL. SampleNet is shown to outperform both $\beta$-NLL and MDN by a wide margin when comparing the ES on both datasets. 

In addition, figures~\ref{fig:weather_qualitative} and~\ref{fig:traffic_qualitative} qualitatively show that SampleNet predicts good approximations of the test set distributions on both datasets. On the Weather dataset, MDN is shown to be able to predict a reasonable estimate of the bimodal data generating distribution. On the Traffic dataset, MDN suffers from mode collapse~\cite{makansi2019overcoming}, where the model uses a single Gaussian distribution to explain the data. This is evident from MDN's high value of the ES on the Traffic dataset in table~\ref{table:toy_final_result}, as well as from its samples failing to model the multiple modes of the Traffic test set distribution in \figref{fig:traffic_qualitative}. $\beta$-NLL exhibits poor performance on both datasets as it is restricted to estimating a unimodal Gaussian distribution, which we visually verify in figures~\ref{fig:weather_qualitative} and~\ref{fig:traffic_qualitative}. Other unimodal parametric models also fail to learn multimodal distributions, we provide their results on the Weather dataset in table~\ref{table:weather_proof_of_baselines} of the appendix. \looseness=-1 

\begin{figure*}[t]
   \centering
     \begin{subfigure}[b]{0.245\textwidth}
    \includegraphics[width=\textwidth]{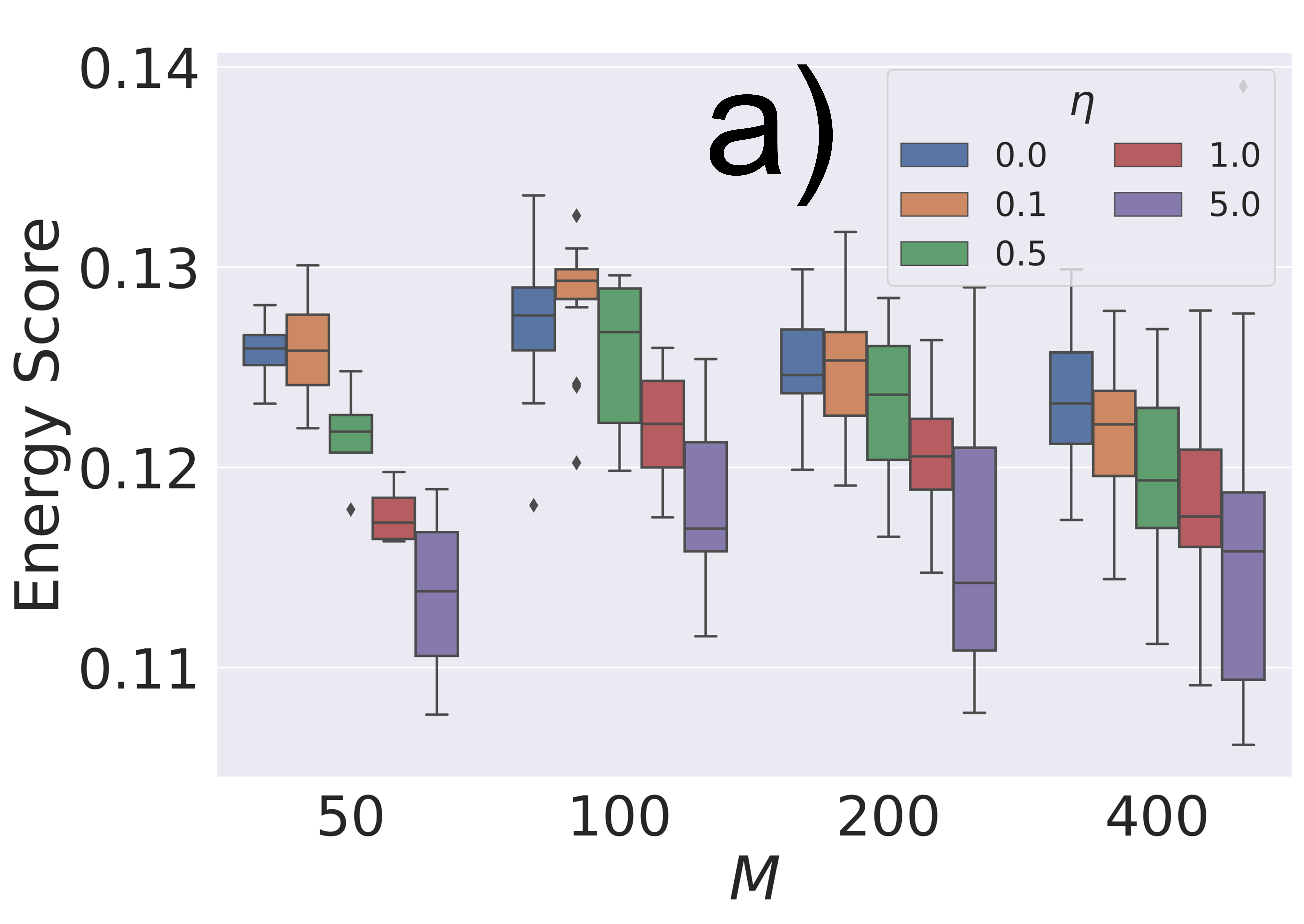}
    \label{fig:ablation_weather_eta}
     \end{subfigure}
    \begin{subfigure}[b]{0.245\textwidth}
         \includegraphics[width=\textwidth]{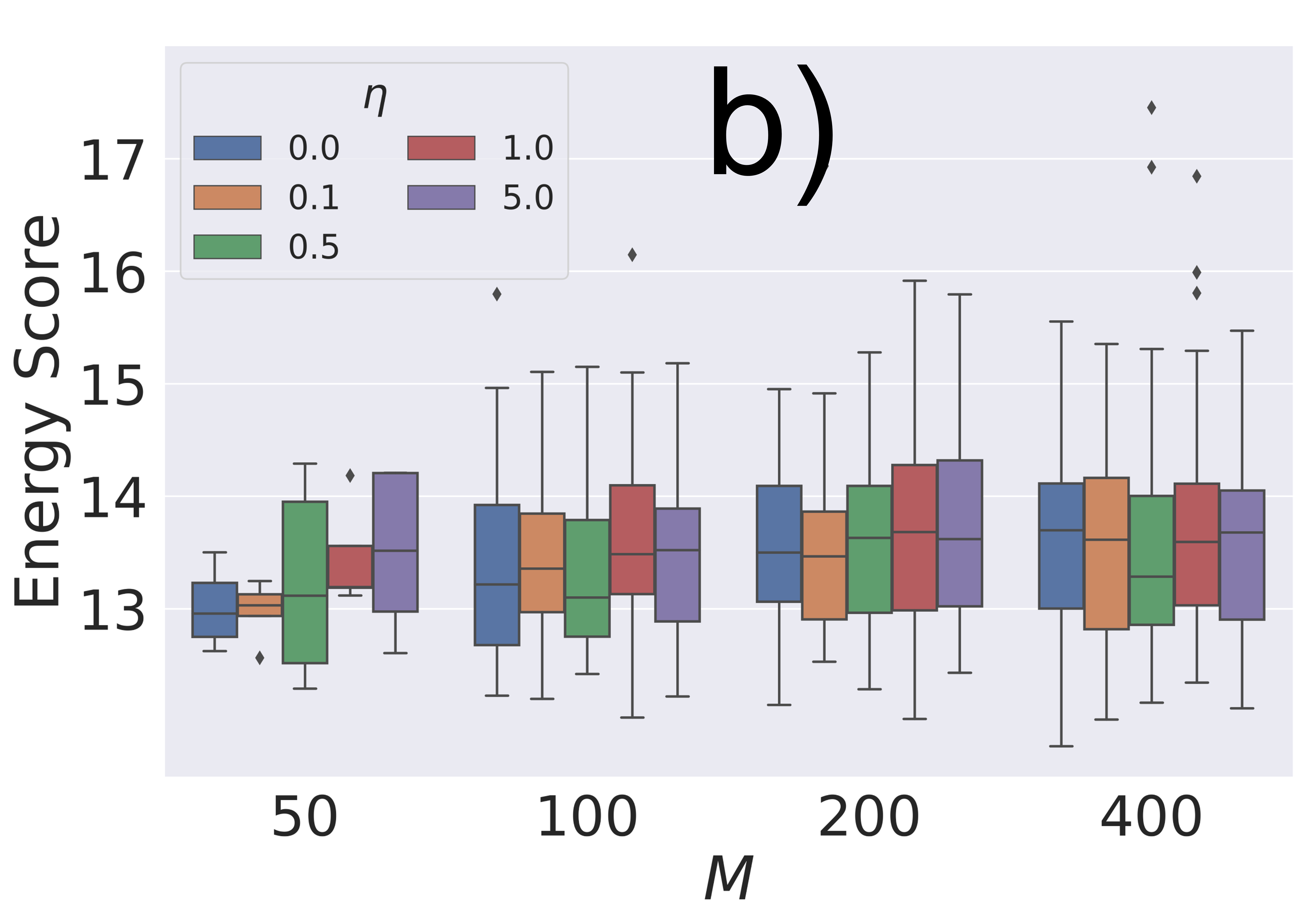}
        \label{fig:ablation_traffic_eta}
     \end{subfigure}
    \begin{subfigure}[b]{0.245\textwidth}
         \includegraphics[width=\textwidth]{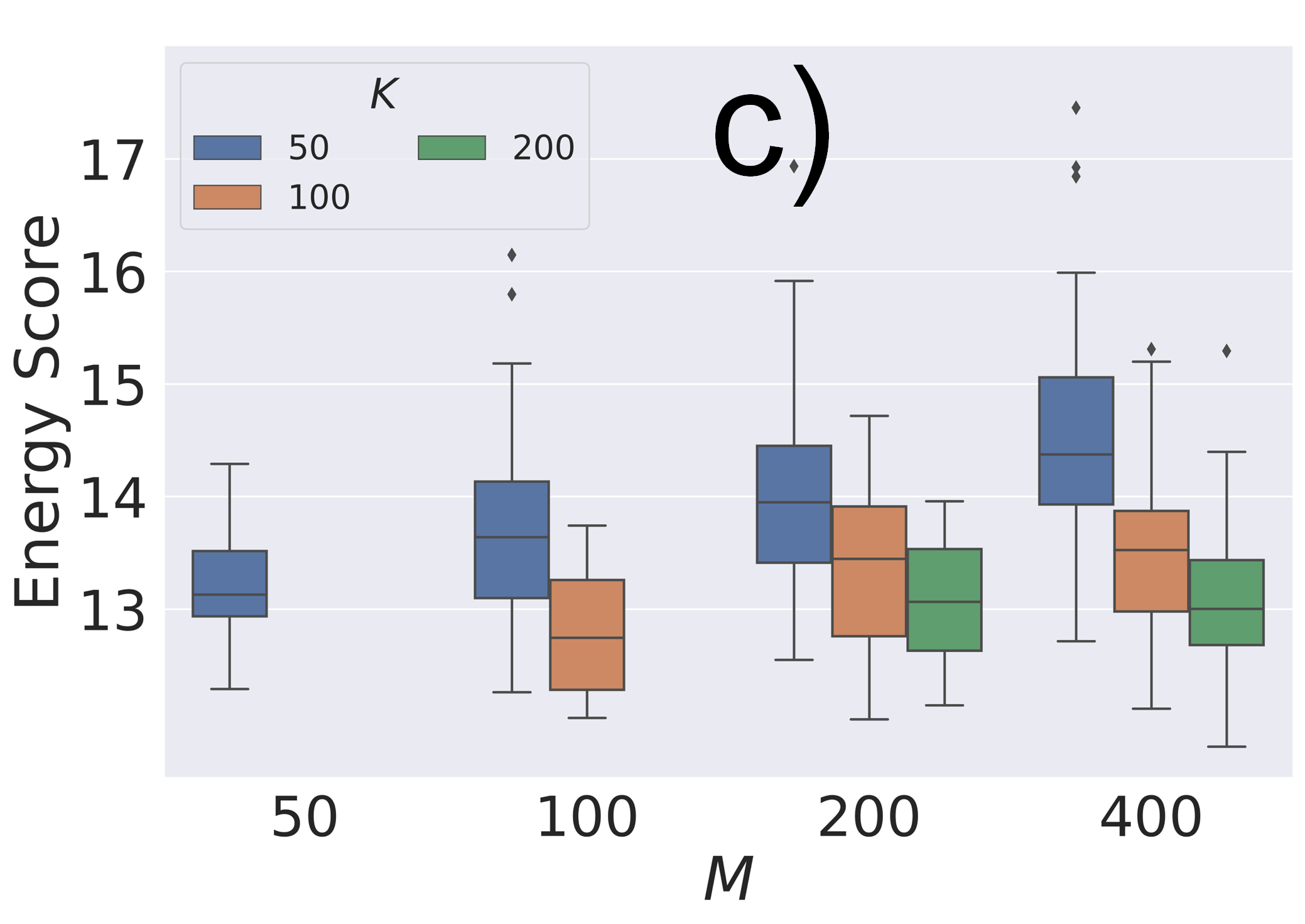}
         \label{fig:ablation_traffic_k}
     \end{subfigure}
    \begin{subfigure}[b]{0.245\textwidth}
    \includegraphics[width=\textwidth]{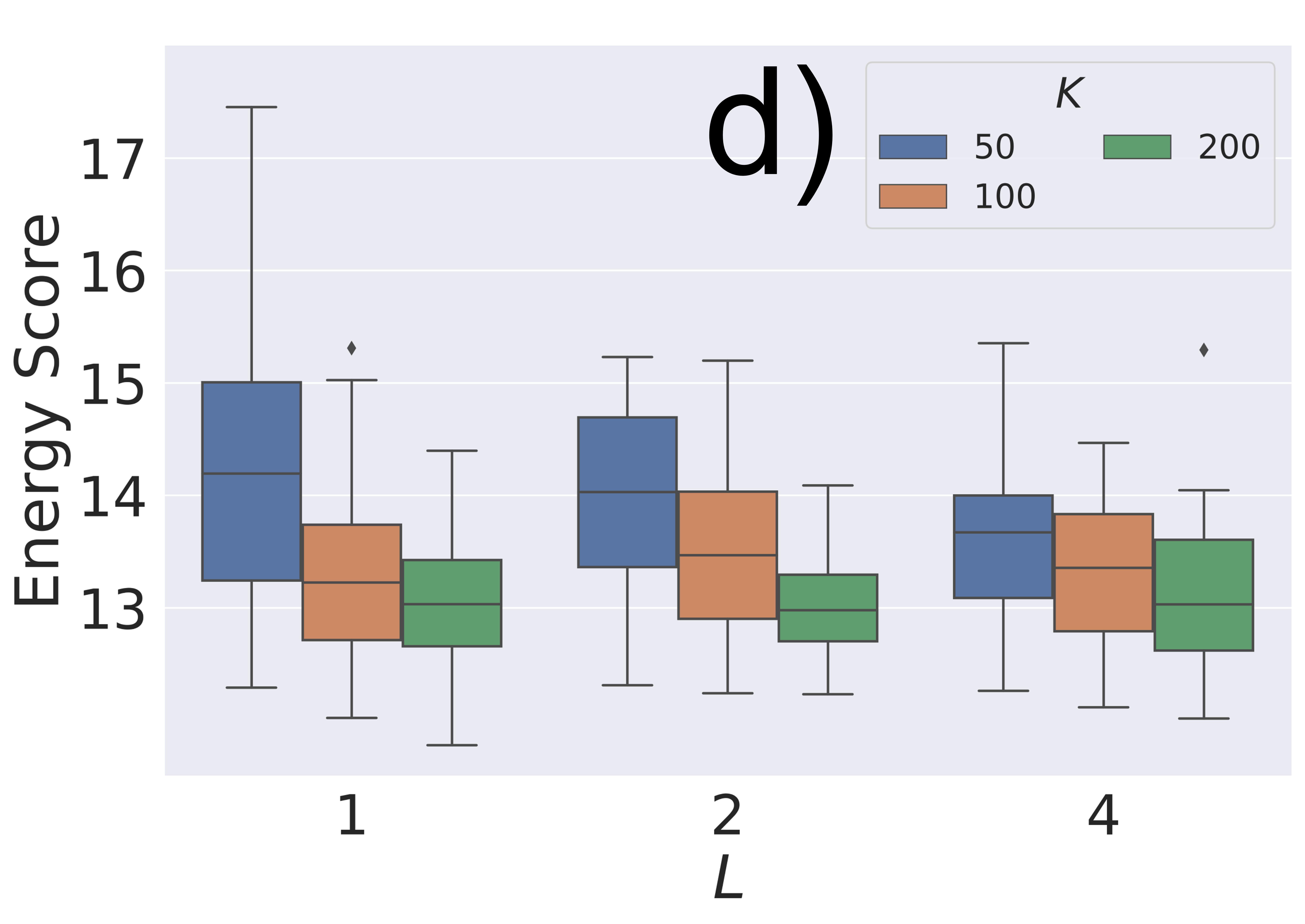}
    \label{fig:ablation_traffic_l}
   \end{subfigure}
   \vspace*{-3mm}
    \caption{Box plots of the ES against: \textbf{a)} $M$ for various values of $\eta$ on the Weather dataset. \textbf{b)} $M$ for various values of $\eta$ on Traffic dataset. \textbf{c)} $M$ for various values of $K$ on the Traffic dataset \textbf{d)} $L$ for various values of $K$ on the Traffic dataset. All experiments are repeated using $5$ random seeds. Additionally, we aggregate over all hyperparameters that are not explicitly considered in each plot. As such, the minimum value of each box plot represents the best achievable performance of SampleNet when using a particular value of plotted hyperparameters, e.g. the lowest ES using a specific value of $\eta$ and $M$ while considering all values of $L$ and $K$ in \textbf{(a)}.}
    \label{fig:all_ablations}
\end{figure*}

\begin{table*}[t]
   \centering
   \caption{The mean and standard deviation of the ES over the train-test splits of the real-world regression datasets. Best performing methods are marked in bold. The $\dagger$ symbol indicates a SampleNet trained without OT regularization ($\eta=0$).}
   \resizebox{\textwidth}{!}{
  \begin{tabular}{lccc cccccc}
    \label{table:energy_score_real_regression}
     Dataset& N & $D_{in}$ & $D_{out}$ & Dropout & $\beta$-NLL & Dropout \& $\beta$-NLL & Student-t & MDN & SampleNet\\
    \cmidrule(lr){1-4} \cmidrule(lr){5-10}
    Boston& 506 & 13 & 1 & $\textbf{1.572}\pm \textbf{0.297}$& $1.938\pm 0.344$  &  $1.732\pm 0.243$ & $\textbf{1.662} \pm \textbf{0.346}$ & $\textbf{1.644}\pm \textbf{0.294}$ & $\textbf{1.559 }\pm\textbf{0.355}$\\
    Kin8nm& 8192 & 8 & 1 & $ 0.056\pm 0.002$& $0.044 \pm  0.002$  &  $ 0.066\pm0.002 $ & $0.048 \pm0.002 $ & $ \textbf{0.041} \pm \textbf{0.001}$ & $ \textbf{0.042 }\pm \textbf{0.001}$\\
    Power& 9568 & 4 & 1 & $ 11.317\pm 0.049$& $ \textbf{2.282}\pm  \textbf{0.061}$  &  $ 10.701\pm 0.206$ & $ \textbf{2.289}\pm \textbf{0.049}$ & $2.325 \pm 0.063$ & $ \textbf{2.263}\pm \textbf{0.056}$\\
    Yacht& 307 & 7 & 1 & $ 0.842 \pm 0.155$& $ 3.718\pm 0.57 $  &  $ 0.692\pm 0.083$ & $ 4.67 \pm 0.834$ & $ 0.597 \pm 0.198$ & $ \textbf{0.179} \pm \textbf{0.056}^\dagger$\\
    Concrete& 1030 & 8 & 1 & $ 3.082\pm 0.257$& $ 3.416\pm 0.267 $  &  $ 3.475\pm 0.221$ & $ 3.301\pm0.255 $ & $ \textbf{3.069}\pm \textbf{0.267}$ & $ \textbf{2.823}\pm \textbf{0.299}^\dagger$\\
    Wine (red)& 1599 & 11 & 1 & $ 0.359\pm 0.019$& $ 0.359\pm 0.020 $  &  $ 0.386\pm0.013 $ & $ \textbf{0.361}\pm \textbf{0.023}$ & $ 0.362\pm 0.020$ & $ \textbf{0.340}\pm \textbf{0.020} $\\
    Wine (white)& 4898 & 11 & 1 & $ 0.397\pm 0.010$& $ 0.398\pm 0.009 $  &  $ 0.427\pm0.009 $ & $ 0.394\pm 0.012$ & $ 0.395\pm 0.010$ & $ \textbf{0.379}\pm \textbf{0.009} $\\
    Naval& 11934 & 16 & 1 & $ 0.038\pm \text{1e-4}$& $ 0.003\pm 0.003 $  &  $ 0.064\pm 0.002$ & $ 0.004\pm0.001 $ & $  0.003\pm 0.003$ & $ \textbf{0.001} \pm \textbf{7e-5}^\dagger$\\
    Superconductivity& 21263 & 81 & 1& $ 6.730\pm 0.205$& $ 6.593\pm 0.294 $  &  $ 6.961\pm 0.115$ & $ 6.597\pm 0.139$ & $ 6.415\pm0.122 $ & $ \textbf{6.035}\pm \textbf{0.127}^\dagger$\\
    Protein& 45730 & 9 & 1 & $ 3.017\pm 0.054$& $ 2.392\pm0.016  $  &  $ 2.612\pm 0.021$ & $ 2.474\pm0.026$ & $ \textbf{2.116}\pm \textbf{0.031}$ & $ \textbf{2.120}\pm \textbf{0.018}$\\
    Year& 515345 & 90 & 1 &$ 48.288\pm 0.058$& $ 4.955\pm 0.264 $  &  $ 66.388\pm 0.155$ & $ 4.784\pm 0.088$ & $ 4.800\pm 0.087$ & $ \textbf{4.606}\pm \textbf{0.017}$\\
    ObjectSlide& 136800 & 5 & 1 & $0.010 \pm $7e-5& \textbf{34e-4}$\pm$\textbf{7e-5} &  0.023$\pm$5e-5 & 47e-4$\pm$6e-5 & \textbf{33e-4}$\pm$\textbf{6e-5} &  \textbf{35e-4}$\pm$ \textbf{3e-5}$^\dagger$\\
    Energy& 768 & 8 & 2 & $ 1.319\pm 0.079$& $ 0.936\pm  0.165$  &  $ 4.331\pm 0.173$ & $ 0.963\pm0.221 $ & $ 1.211\pm0.223 $ & $\textbf{ 0.729}\pm \textbf{0.068}$\\
    Carbon& 10721 & 5 & 3 & $ 0.027\pm 0.028$& $ 0.061\pm  0.005$  &  $ 0.001\pm  0.032$ & $ 0.006 \pm 0.001$ & $ \textbf{0.004}\pm \textbf{0.001}$ & $\textbf{0.004} \pm \textbf{0.001}$\\
    Fetch-Pick\&Place& 638400 & 29 & 3 & 35e-4$ \pm $ 1e-5& 24e-4$ \pm  $ 1e-5 &  430e-4$ \pm $6e-5 & 27e-4$ \pm $9e-5 & 35e-4$ \pm $2e-5 & \textbf{20e-4}$\pm $\textbf{2e-5}$^\dagger$\\
    \midrule
    \midrule    
    Total Top Performance& - & - & - & 1& 2 & 0  & 3& 6& \textbf{15}\\
    \bottomrule
   \end{tabular}}
\end{table*}
\begin{table*}[t]
   \centering
     \caption{The mean and standard deviation of the Gaussian NLL over the train-test splits of the real-world regression datasets. Best performing methods are marked in bold. For computing the Gaussian NLL for SampleNet, we use the empirical mean and variance of the predicted samples.}
   \resizebox{\textwidth}{!}{
  \begin{tabular}{lccc cccccc}
    \label{table:nll_real_regression}
     Dataset& N & $D_{in}$ & $D_{out}$ & Dropout & $\beta$-NLL & Dropout \& $\beta$-NLL & Student-t & MDN & SampleNet\\
    \cmidrule(lr){1-4} \cmidrule(lr){5-10}
    Boston& 506 & 13 & 1 & $\textbf{2.634}\pm \textbf{0.353}$& $2.768\pm 0.325$  &  $2.596\pm 0.103$ & $\textbf{2.589} \pm \textbf{0.362}$ & $\textbf{2.781}\pm \textbf{0.651}$ & $\textbf{2.413 }\pm\textbf{0.273}$\\
    Kin8nm& 8192 & 8 & 1 & $ -0.795\pm 0.057$& $-1.174 \pm  0.035$  &  $ -0.615\pm0.027 $ & $-1.087 \pm0.038 $ & $ -\textbf{1.242} \pm \textbf{0.028}$ & $-1.213\pm 0.025$\\
    Power& 9568 & 4 & 1 & $ 4.751\pm 0.002$& $ \textbf{2.839}\pm  \textbf{0.048}$  &  $ 4.494\pm 0.023$ & $ \textbf{2.833}\pm \textbf{0.037}$ & $\textbf{2.859} \pm \textbf{0.038}$ & $ \textbf{2.851}\pm \textbf{0.037}$\\
    Yacht& 307 & 7 & 1 & $ 1.799\pm  0.113$& $ 3.465\pm  0.258 $  &  $ 1.676\pm 0.113$ & $ 5.692 \pm 2.59$ & $ \textbf{0.214} \pm \textbf{0.423}$ & $ \textbf{0.033} \pm \textbf{0.188}^\dagger$\\
    Concrete& 1030 & 8 & 1 & $ 3.392\pm 0.305$& $ 3.314\pm 0.144 $  &  $ 3.227\pm 0.072$ & $ 3.172\pm0.122 $ & $ \textbf{3.320}\pm \textbf{0.589}$ & $ \textbf{3.060}\pm \textbf{0.138}^\dagger$\\
    Wine (red)& 1599 & 11 & 1 & $ \textbf{0.980}\pm \textbf{0.062}$& $\textbf{0.992}\pm \textbf{0.130} $  &  $ 1.117\pm0.027  $ & $ 156.548\pm 386.887$ & $ 1.226\pm 0.806$ & $ \textbf{0.945}\pm \textbf{0.068}$\\
    Wine (white)& 4898 & 11 & 1 & $ 1.109\pm 0.038$& $ 1.076\pm 0.033 $  &  $ 1.214\pm0.016 $ & $ \textbf{1.058}\pm \textbf{0.026}$ & $ 2.560\pm 6.358$ & $ \textbf{1.045}\pm \textbf{0.024} $\\
    Naval& 11934 & 16 & 1 & $-1.819\pm 0.071$& $ -\textbf{10.045}\pm \textbf{0.481} $  &  $ -1.819\pm 0.071$ & $ -7.873\pm0.659  $ & $  -9.974\pm 0.401$ & $ -\textbf{10.224} \pm \textbf{0.146}^\dagger$\\
    Superconductivity& 21263 & 81 & 1& $ 5.501\pm 0.238$& $ \textbf{3.721}\pm \textbf{0.197} $  &  $ 3.771\pm 0.034$ & $4.260\pm 2.465$ & $ 3.983\pm0.443  $ & $ \textbf{3.602}\pm \textbf{0.041}^\dagger$\\
    Protein& 45730 & 9 & 1 & $ 10.928\pm 0.208$& $ 3.105\pm 0.313 $  &  $ \textbf{2.904}\pm\textbf{0.008} $ & $ 4.996\pm 2.640$ & $ \textbf{2.912}\pm \textbf{0.106}$ & $\textbf{ 2.845}\pm \textbf{0.007} $\\
    Year& 515345 & 90 & 1 &$ 6.197\pm 0.001$& $ \textbf{3.510}\pm \textbf{0.060}$  &  $ 6.500\pm 0.002$ & $ 1496.605\pm 1308.447 $ & $ \textbf{3.579}\pm \textbf{0.047}$ & $ \textbf{3.531}\pm \textbf{0.005}$\\
    ObjectSlide& 136800 & 5 & 1 & $ -2.287 \pm 0.148$& $ -\textbf{4.996}\pm \textbf{1.015} $  &  $- 1.857\pm0.049 $ & $ -4.010\pm0.347$ & $-3.461 \pm2.054 $ & $ -3.016\pm 0.309^\dagger$\\
    Energy& 768 & 8 & 2 & $  3.763\pm 0.070$& $ 2.323\pm  0.390$  &  $ 6.389\pm 0.063$ & $ 2.450\pm0.562 $ & $ 2.844\pm1.239 $ & $\textbf{2.001}\pm \textbf{0.299}$\\
    Carbon& 10721 & 5 & 3 & $ -3.967\pm 0.043$& $ -\textbf{6.854}\pm  \textbf{9.193}$  &  $ -6.737\pm  0.047$ & $ -\textbf{10.589} \pm \textbf{0.856}$ & $ 39.275\pm 47.959$ & $-7.461 \pm 3.283$\\
    Fetch-Pick\&Place& 638400 & 29 & 3 & $-12.703 \pm 0.337$& $-14.613 \pm 0.383 $  &  $-6.845 \pm 0.107$ & $ -14.078\pm 0.184$ & $ -13.092\pm 0.372$ & $ -\textbf{14.800}\pm \textbf{0.088}^\dagger$\\
    \midrule
    \midrule
    Total Top Performance& - & - & - & 2& 7 & 1  & 4& 7& \textbf{12}\\
    \bottomrule
   \end{tabular}}
\end{table*}

\subsection{Real-world Regression Datasets}
We evaluate the performance of SampleNet in comparison to baselines on datasets from the UCI regression benchmark~\cite{uci}. We use $13$ datasets that have been used in the evaluation of recent methods from literature such as $\beta$-NLL~\cite{seitzer2022on} and Student-t regression~\cite{skafte2019reliable}. Following~\cite{seitzer2022on}, we also add $2$ robot dynamics models datasets, ObjectSlide~\cite{seitzer2021causal} and Fetch-PickAndPlace~\cite{seitzer2022on}, to our evaluation for a total of $15$ datasets. The input dimension $D_{in}$, output dimension $D_{out}$, and number of instances $N$ for the $15$ datasets can be found in table~\ref{table:energy_score_real_regression}. For all methods, we report the mean and standard deviation of the ES in table~\ref{table:energy_score_real_regression} and the Gaussian NLL in table~\ref{table:nll_real_regression} computed over 20 train-test splits, similar to~\cite{seitzer2022on}. We report values for the RMSE in table~\ref{table:rmse_real_regression} of the appendix. We also perform a two-sided Kolmogorov-Smirnov test with a $p\leq0.05$ significance level and mark in bold any methods that produce statistically indistinguishable results when compared to the best performing method in tables~\ref{table:energy_score_real_regression},~\ref{table:nll_real_regression}, and~\ref{table:rmse_real_regression}. More details on the datasets and hyperparameters can be found in \secref{appendix:regression_datasets}.\looseness=-1

Table~\ref{table:rmse_real_regression} shows that SampleNet as well as most other baselines perform equally well when measuring their RMSE, meaning that performance improvements on proper scoring rules can be attributed to improvements in estimating the predictive distribution. When aggregating results over all $15$ datasets, SampleNet exhibits performance that is better or on par with other baselines on $15$ out of $15$ datasets when using the ES for evaluation (last row of table~\ref{table:energy_score_real_regression}) and $12$ out of $15$ datasets when using the Gaussian NLL for evaluation(last row of table~\ref{table:nll_real_regression}). In addition, SampleNet results in a statistically significant improvement in ES on $8$ out of $15$ datasets and a statisticalky significant improvement in Gaussian NLL on $2$ out of $15$ datasets when compared to the rest of our baselines.

\subsection{Hyperparameter Sensitivity Analysis}
One disadvantage of SampleNet is the introduction of $4$ additional hyperparameters to optimize for: the predicted sample size $M$, the minibatch loss sample size $K$, the minibatch loss number of repetitions $L$, the OT regularization strength parameter $\eta$. Using the Weather and Traffic datasets, we perform a sensitivity analysis to understand the impact of varying these hyperparameters by training SampleNets over all combinations of $M = \{50, 100, 200, 400\}$, $K=\{50, 100, 200\}$, $L=\{1, 2, 3, 4\}$, and $\eta=\{0.0, 0.1, 0.5, 1.0, 5.0\}$. We use a Gaussian prior for regularization.

Figures~\ref{fig:all_ablations}a~and~\ref{fig:all_ablations}b show the impact of varying $\eta$ on the ES for various values of $M$ on the Weather and Traffic datasets, respectively, aggregated over all values of $L$ and $K$. On the Traffic dataset, $\eta$ is shown to have a minimal impact on performance. On the other hand, results on the Weather dataset show that increasing $\eta$ results in a large decrease in the ES for all values of $M$. When noting that the Traffic dataset contains $6$ times the amount of training data when compared to the Weather dataset, our proposed regularization improves the results on the smaller dataset. In addition, the Weather dataset is originally generated from two unimodal Gaussian distributions, which could benefit from our regularization loss driving the solution to optimally transport a Gaussian prior. We notice similar sensitivity to the value of $\eta$ when testing on real-world datasets in Table~\ref{table:energy_score_real_regression}, where we find no consistency in the value of $\eta$ or the choice of prior among the $15$ datasets when looking at the best performing SampleNet configuration.

\Figref{fig:all_ablations}c shows the impact of varying $K$ on the ES for various values of $M$ on Traffic datasets, aggregated over all values of $L$ and $\eta$. For a fixed value of $M$, increasing $K$ results in an improvement in the accuracy of predicted distributions as the minibatch losses in ~\eqref{eq:minibatch_energy_score} and ~\eqref{eq:minibatch_sinkhorn} become a more accurate approximation of the original losses in ~\eqref{eq:energy_score_samples} and~\eqref{eq:sinkhorn_divergence}. The results in \figref{fig:all_ablations}d show a similar behavior, where the ES decreases when increasing the number of repetitions $L$ in ~\eqref{eq:minibatch_energy_score} and ~\eqref{eq:minibatch_sinkhorn} for a fixed value of $K$ on the Traffic datasets. Similar results on the Weather dataset can be found in \figref{fig:ablations_weather_additional} of the appendix. 
The most important observation from \figref{fig:all_ablations} is that the minimum achievable ES (lower point of the box plots) does not substantially decrease as we increase the number of output samples $M$, meaning that we can achieve similar performance using any value of $M$ and enough hyperparameter optimization. A similar behavior is observed on UCI and dynamics models datasets, where we notice that the best performing SampleNet configurations in table~\ref{table:samplenet_regression_hyperparameters} used $M<200$ for $11$ out of $15$ datasets. These observations alleviate a major concern about the dependence of our sample-based approach on a large output sample size to achieve good performance.
\begin{table*}[htb]
\centering
\caption{Results on the NYUv2 dataset. SampleNet trained with ($\eta=0$) is marked with $\dagger$.}
\label{table:depth_prediction_results}
\resizebox{\textwidth}{!}{
\begin{tabular}{l ccccccccc cc cc}
& \multicolumn{9}{c}{Deterministic Metrics} & \multicolumn{2}{c}{Scoring Rules}& \multicolumn{2}{c}{Average Var} \\
 Method & $\delta$1↑   & $\delta$2↑  & $\delta$3↑  & REL↓     & Sq Rel↓   & RMS↓     & RMS log↓   & log10↓  & SI log↓          & NLL↓            & ES↓ &  All & Border\\ \cmidrule(lr){1-1}\cmidrule(lr){2-10}\cmidrule(lr){11-12}\cmidrule(lr){13-14}
$\beta$-NLL~\cite{seitzer2022on}        & 0.8878          & 0.9798          & 0.9951          & 0.1099          & 0.0666          & 0.3825          & 0.1390          & 0.0462          & 11.346          & 7.237 $\pm$ 6.85e-1          & 0.2296 $\pm$ 1.43e-3    & 0.0304 &   0.0533    \\
Student-t~\cite{skafte2019reliable}    & 0.8844          & 0.9794          & 0.9948          & 0.1101          & 0.0656          & 0.3868          & 0.1406          & 0.0464          & 11.494        & 7.187 $\pm$ 1.56e-1          & \textbf{0.2274} $\pm$ 7.09e-4 & 0.0410 & 0.0731\\ 
MDN~\cite{bishop1994mixture}   & 0.8812          & 0.9792          & 0.9951          & 0.1114          & 0.0679          & 0.3922          & 0.1425          & 0.0467          & 11.722          & 8.380 $\pm$ 9.59e-1          & 0.2321 $\pm$ 2.72e-4      &0.0445 & 0.0749     \\ \cmidrule(lr){1-1}\cmidrule(lr){2-10}\cmidrule(lr){11-12}\cmidrule(lr){13-14}

SampleNet$^\dagger$    & 0.8891          & 0.9800          & 0.9950          & 0.1071          & 0.0642          & 0.3795          & 0.1383          & 0.0456          & 11.332          & \textbf{1.397} $\pm$ 7.05e-4 & 0.2371 $\pm$ 5.43e-4  & 0.0157& 0.0340 \\
SampleNet & \textbf{0.8900} & \textbf{0.9820} & \textbf{0.9959} & \textbf{0.1066} & \textbf{0.0601} & \textbf{0.3757} & \textbf{0.1366} & \textbf{0.0451} & \textbf{11.195} & \textbf{1.398} $\pm$ 1.22e-3 & 0.2354 $\pm$ 4.01e-3 & 0.0115  & 0.0261   \\ \bottomrule
\end{tabular}}
\end{table*}
When applying SampleNet to a new problem, our sensitivity analysis indicates that the best approach is to choose the highest value of $M$ that fits the computational memory and time constraints with $K$=$M$ and $L$=$1$. Hyperparameter optimization can then be performed for $\eta$, the choice of prior, as well as any other standard neural network hyperparameters such as the learning rate. Our minibatch subsampling losses leave the option for practitioners to increase the number of output samples $M$ if needed while maintaining adequate performance.

\subsection{Monocular Depth Prediction}
Following our own recommendations on setting hyperparameters, we use SampleNet to estimate predictive distributions on the large-scale  computer vision task of monocular depth prediction. We use the NYUv2 dataset~\cite{nyuv2} for training and testing. We modify the same base depth prediction architecture used by $\beta$-NLL in~\cite{seitzer2022on} to generate samples and train using the loss in~\ref{eq:total_loss}. We compare the performance of SampleNet on deterministic metrics and proper scoring rules to three baselines $\beta$-NLL, Student-t regression, and MDN. More details about the experimental setup and metrics can be found in \secref{appendix:depth_prediction_datasets}.\looseness=-1

Our compute resources allow us to set a maximum value of $M=25$ when $K$=$M$ and $L$=1 before running out of memory. For these values of $M$, $L$, and $K$, we train SampleNets with $\eta=\{0.0, 0.1, 0.5, 1.0\}$ using a Gaussian prior, and show the results for the best performing configuration ($\eta=1.0$) as well as the unregularized variant ($\eta=0.0$) in Table~\ref{table:depth_prediction_results}. Using SampleNet leads to a significant reduction in Gaussian NLL on the test set when compared to other baselines while maintaining a comparable value of the ES, mainly because SampleNet generates much sharper distributions. The sharpness property can be quantitatively shown in the last two columns of table~\ref{table:depth_prediction_results}, where distributions predicted from SampleNet are shown to have a much lower average variance when compared to those from baselines. Qualitative evidence of sharper distributions is also provided in \figref{fig:depth_pc} of the appendix. SampleNet is also shown to outperform all baselines on standard deterministic depth prediction metrics, meaning that any improvement in predictive distributions does not come at the cost of accuracy on the original depth prediction problem.

Our findings align well with those in previous literature, where Gaussian parametric distributions predicted from neural networks are found to have high variance if learned on high variance training data using the NLL~\cite{harakeh2021estimating}. Due to discontinuities of depth values at object boundaries in images, we hypothesize that the baselines learn high variance distributions to accommodate for their inability to model multiple depth modes. SampleNet seems to resolve this issue, with the regularized configuration leading to even sharper distributions when compared to the unregularized configuration (last two rows of table~\ref{table:depth_prediction_results}).

\section{Conclusion}
\label{sec:conclusion}
We present SampleNet, a scalable approach to model uncertainty in deep neural networks without imposing a parametric form on the predictive distribution. Our simple solution provides an easy-to-implement tool for enabling the estimation of regression predictive distributions on real-world problems. One limitation of SampleNet is only accurately predicting the uncertainty on in-distribution data. Modifying SampleNet's regularization to output high uncertainty values on out-of-distribution data (e.g. as in noise-contrastive priors~\cite{hafner2020noise}) would be an interesting avenue for future work. Example code can be found at: https://samplenet.github.io/.
\section*{Acknowledgements}
A. H. was supported by the DEEL project and the IVADO postdoctoral funding. L. P. is supported by
the Canada CIFAR AI Chairs Program. The work was also supported by the National Science and Engineering Research Council of Canada under the Discovery Grant Program.

% Use \bibliography{yourbibfile} instead or the References section will not appear in your paper
\bibliography{aaai23}

%%%%%%%%%%%%
% Appendix %
%%%%%%%%%%%%
\appendix
\newpage
\clearpage
\counterwithin{figure}{section}
\counterwithin{table}{section}
\section{Experimental Setup}
\label{appendix:experimental_setup}

\subsection{Implementation Details}
All methods share the same neural network and only differ in how we represent the output probability distribution through the last linear layers. All methods are trained with Adam optimizer using the default settings in PyTorch. Whenever we output a value for variance, we apply SoftPlus function to produce positive values. If the output dimension is $d > 1$, we assume a fully factored Gaussian distribution with a diagonal covariance matrix. For SampleNet and MDN, we compute the mean and variance from predicted samples when those statistics are needed for evaluation e.g. when the Gaussian NLL in~\ref{eq:negative_log_likelihood}. For all models, we perform a grid search over various hyperparameters to determine the best performing configuration for every dataset.

\textbf{$\beta$-NLL}: We follow the implementation of Variance Networks proposed in~\cite{seitzer2022on} to predict a single mean and variance for each input $\vx_n$. We train models using the $\beta$-NLL loss:
\begin{equation*}
    \mathcal{L}_{\beta\text{-NLL}}(\vtheta) = -\frac{1}{N}\sum\limits_{n=1}^N (\sigma^2_\theta(\vx_n)^\beta)\log p_\vtheta(\vy_n|\vx_n),
\label{eq:beta_negative_log_likelihood}
\end{equation*}
where $\sigma^2_\theta(\vx_n)$ is the predicted variance (with gradient stop operation) and $\beta$ is treated as a hyperparameter. For all experiments, we perform a grid search over $\beta = [0.0, 0.25, 0.5, 0.75, 1.0]$ to choose the best performing method.

\textbf{Ensemble}~\cite{lakshminarayanan2017simple}: We create an ensemble of $5$ $\beta$-NLL models. During inference, the mean and variance are approximated according to an equally weighted Gaussian mixture model as proposed in~\cite{lakshminarayanan2017simple}. \textit{Ensembles were not shown to provide a substantial improvement over a single $\beta$-NLL model}. We can also create ensembles of SampleNets, MDNs, or any other model, making a comparison of these methods agains multiple $\beta$-NLL neural networks unfair. As such, we do not consider ensembles as a baseline in our experiments. For all experiments, we perform a grid search over $\beta = [0.0, 0.25, 0.5, 0.75, 1.0]$ to choose the best performing method.

\textbf{Mixture Density Networks (MDNs)}: We implement MDN following~\cite{bishop1994mixture} using \href{https://github.com/hardmaru/pytorch\_notebooks/blob/master/mixture\_density\_networks.ipynb}{code at this link} as a reference. Our implementation predicts $J$ mean and variance estimates as well as a mixture weight vector $W \in \R^J$ for every input $\vx_n$. During inference, we generate samples from the mixture components to be used for evaluation. $J$ is treated as a hyperparameter. For all experiments, we perform a grid search over $J = [2,3,4,5]$ to choose the best performing method.

\textbf{Dropout}: We apply dropout during training and inference using a dropout probability $p_{drop}$, which we treat as a hyperparameter.  For all experiments, we perform a grid search over $p_{drop} = [0.01, 0.05, 0.1]$ to choose the best performing method.

\textbf{Dropout + $\beta$-NLL}: We apply dropout during training and inference to $\beta-NLL$ variance network with $p_{drop}$ as a hyperparameter. During inference, the mean and variance are approximated according to an equally weighted Gaussian mixture model according to~\cite{lakshminarayanan2017simple}. For all experiments, we perform a grid search over a combination of $\beta = [0.0, 0.25, 0.5, 0.75, 1.0]$ and $p_{drop} = [0.01, 0.05, 0.1]$ to choose the best performing method.

\textbf{Student-t regression}: We follow the implementation of Student-t regression in~\cite{skafte2019reliable} by predicting a gamma distribution from which we sample variance values. The implementation can be found at \href{https://github.com/JorgensenMart/ReliableVarianceNetworks}{at this link}.

\textbf{DiscoNet}~\cite{bouchacourt2016disco}: We attempt to adapt the implementation of DiscoNet found at \href{https://github.com/oval-group/DISCONets}{at this link} to our datasets. The adapted implementation fails to produce meaningful results even on small toy datasets (\figref{fig:toy_on_sinusoidal}).

\subsubsection{SampleNet}
\label{appendix:samplenet}
For our proposed architecture, we simply modify the output linear layer to produce $M \times d$ samples instead of a single value. We train the model using efficient and scalable GPU implementations of the ES and the Sinkhorn Divergence from the geomloss\footnote{\url{https://github.com/jeanfeydy/geomloss}}~\cite{feydy2019interpolating} library.

We decouple the choice of the prior from the specific problem at hand by setting $p_{ot}(\tilde{\rvy}|\vx_n)$ to either be a multivariate Uniform distribution $U((0,1)^d)$ or be an isotropic standard multivariate Gaussian distribution $\mathcal{N}(0,\mI)$. Before computing the Sinkhorn Divergence, we normalize each dimension $\{1, \dots, d\}$ of our predicted samples $\hat{\rvy}$ to be between $0$ and $1$ if our prior is uniform, or standardize the samples to have $0$ mean and unit variance if the prior is Gaussian. Since our samples are constrained to be between $\{-1, 1\}$ when computing the Sinkhorn Divergence in \eqref{eq:sinkhorn_divergence}, we use the Wasserstien-2 distance with $\epsilon=0.05^2$, as recommended by the geomloss library. 

We implement minibatch subsampling following the description~\cite{fatras2019learning}. We treat $M$, $K$, and $L$ in equations~\ref{eq:minibatch_energy_score} and~\ref{eq:minibatch_sinkhorn}, as well as $\eta$ in \eqref{eq:total_loss} as hyperparameters. For all experiments, we perform a grid search over a combination of $M = [50, 100, 200, 400]$, $K=[50, 100, 200]$, $L=[1, 2, 3, 4]$, and $\eta=[0.0, 0.1, 0.5, 1.0, 5.0]$ to choose the best performing method.

%%%%%%%%%%%%%%%%%%%%%%%%%
% Sinusoidal Toy Example%
%%%%%%%%%%%%%%%%%%%%%%%%%
\subsection{Sinusoidal with Heteroscedastic Noise}
\label{appendix:toy_data}
The unimodal version of this synthetic toy example was introduced in~\cite{skafte2019reliable}. $500$ training points are sampled uniformaly in the interval $[0, 10]$ from the function $y = x\sin(x) +  \epsilon_1 x + \epsilon_2$  where $\epsilon_1, \epsilon_2$ are independent samples from a normal distribution with $\mu=0$ and $\sigma=0.3$. Red synthetic outliers in ~\figref{fig:eta_effect} are generated as $y=x+7$. We also create a multimodal version of this toy example by sampling from two functions $y = \cos(x) +  \epsilon_1 x + \epsilon_2 - 5$ and $y = (\epsilon_1+1) x + \epsilon_2 + 5$. On these two toy datasets, all models use $50$ neurons with $\tanh$ activation and are trained with a learning rate of 1e-2 for $2500$ steps using the adam optimizer.

\begin{figure*}[t]
     \begin{subfigure}[b]{0.245\textwidth}
         \includegraphics[trim={5 5 5 5}, clip,width=\textwidth]{png/appendix/toy/varnet.png}
         \caption{$\beta$-NLL}
     \end{subfigure}
          \begin{subfigure}[b]{0.245\textwidth}
         \includegraphics[trim={5 5 5 5}, clip,width=\textwidth]{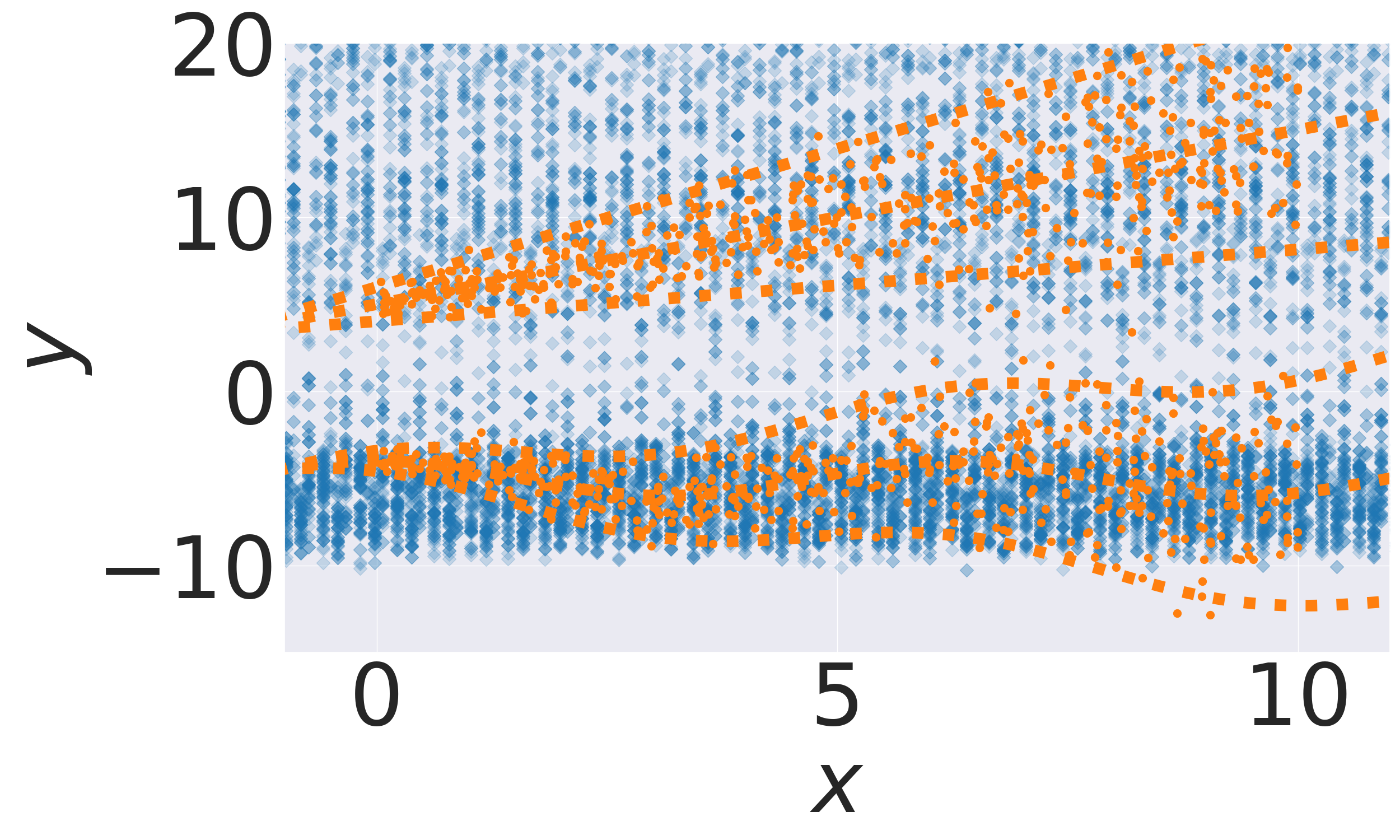}
         \caption{DiscoNet}
         \label{fig:disconet}
     \end{subfigure}
          \begin{subfigure}[b]{0.245\textwidth}
         \includegraphics[trim={5 5 5 5}, clip,width=\textwidth]{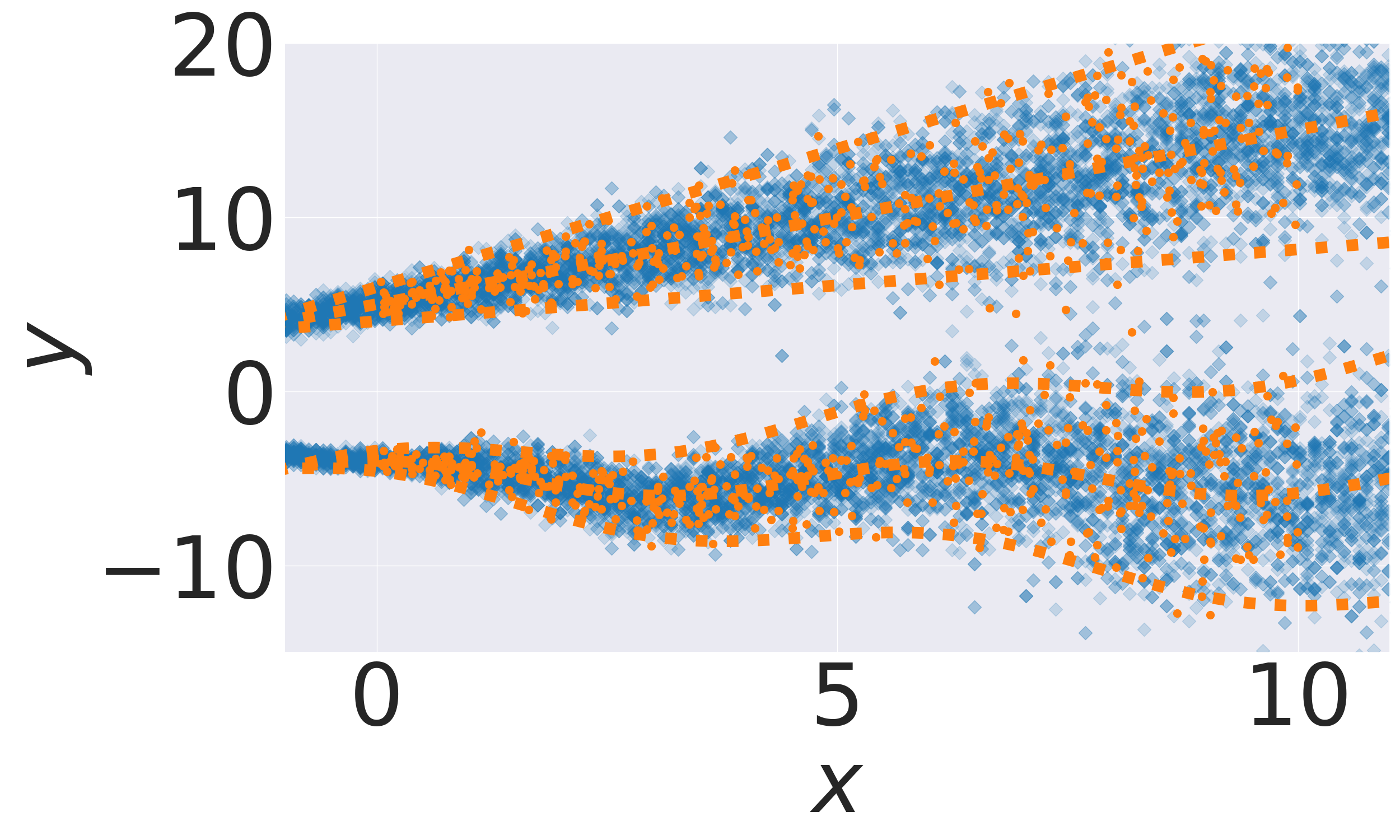}
         \caption{MDN}
     \end{subfigure}
          \begin{subfigure}[b]{0.245\textwidth}
         \includegraphics[trim={5 5 5 5}, clip,width=\textwidth]{png/proper_regularization.png}
         \caption{SampleNet}
     \end{subfigure}
      \caption{Qualitative results of SampleNet and baselines on a multimodal toy example.}
     \label{fig:toy_on_sinusoidal}
\end{figure*}

This multimodal dataset is used as a test to determine if a baseline can model multimodal distributions. For instance, we could not get an implementation of DiscoNet~\cite{bouchacourt2016disco} to converge on the toy examples, so we do not include this method as baseline on our real-world datasets. Multimodal samples generated from DiscoNet are shown in \figref{fig:disconet}.

%%%%%%%%%%%%%%%%%%%%%%
% Multimodal Datasets%
%%%%%%%%%%%%%%%%%%%%%%
\subsection{Multimodal Datasets}
\label{appendix:multimodal_datasets}

\textbf{Weather Dataset}: The ``Weather'' dataset requires models to predict the distribution of the maximum temperature given the day of the year. Maximum daily temperature data is collected from two weather stations in Gardiner, Maine, USA and Arcadia, Florida, USA between the years of 1910 to 2021 using the cli-MATE tool\footnote{\url{https://mrcc.purdue.edu/CLIMATE}}, which is publicly available for research use. The combination of the maximum daily temperatures from the two stations follows a multimodal distribution (\figref{fig:weather_qualitative}). For training, we use $365\times2$ data points collected from both stations during the year 1995. For testing, we use $365\times110$ data points from the remaining 110 years between 1910 and 2021 after excluding the training year, 1995. 

\begin{figure*}[t]
     \begin{subfigure}[b]{0.245\textwidth}
         \includegraphics[trim={5 5 5 5}, clip,width=\textwidth]{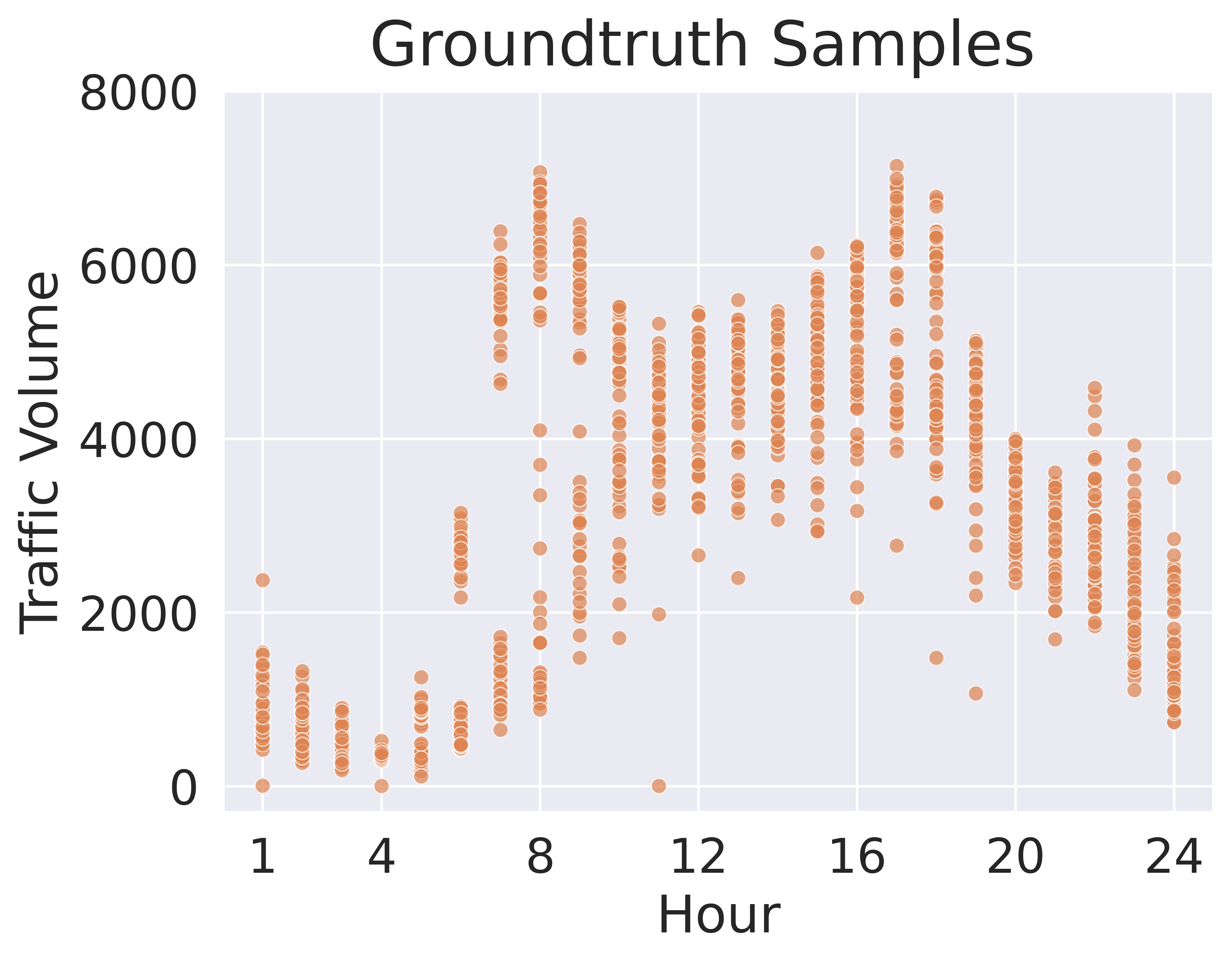}
     \end{subfigure}
          \begin{subfigure}[b]{0.245\textwidth}
         \includegraphics[trim={5 5 5 5}, clip,width=\textwidth]{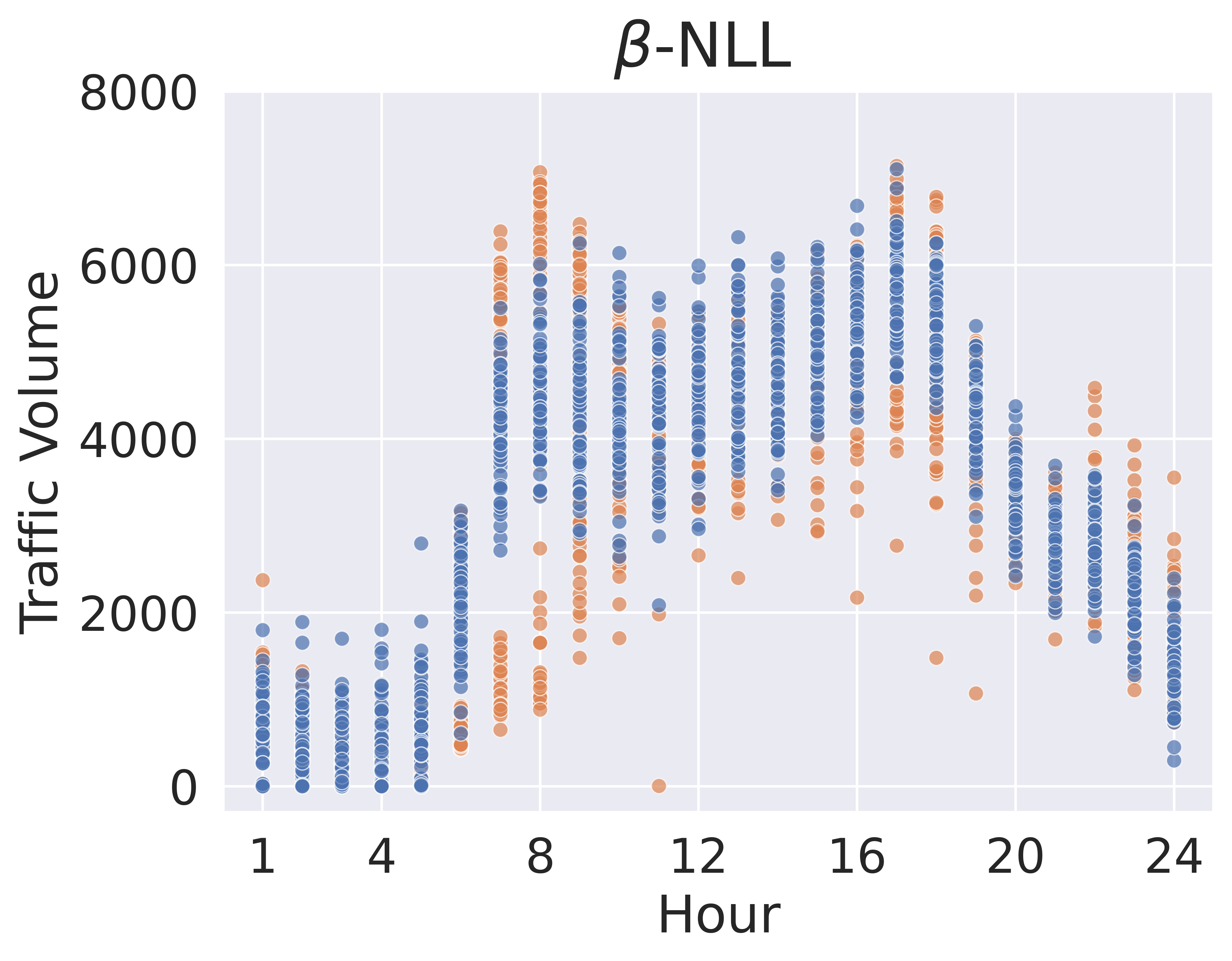}
     \end{subfigure}
          \begin{subfigure}[b]{0.245\textwidth}
         \includegraphics[trim={5 5 5 5}, clip,width=\textwidth]{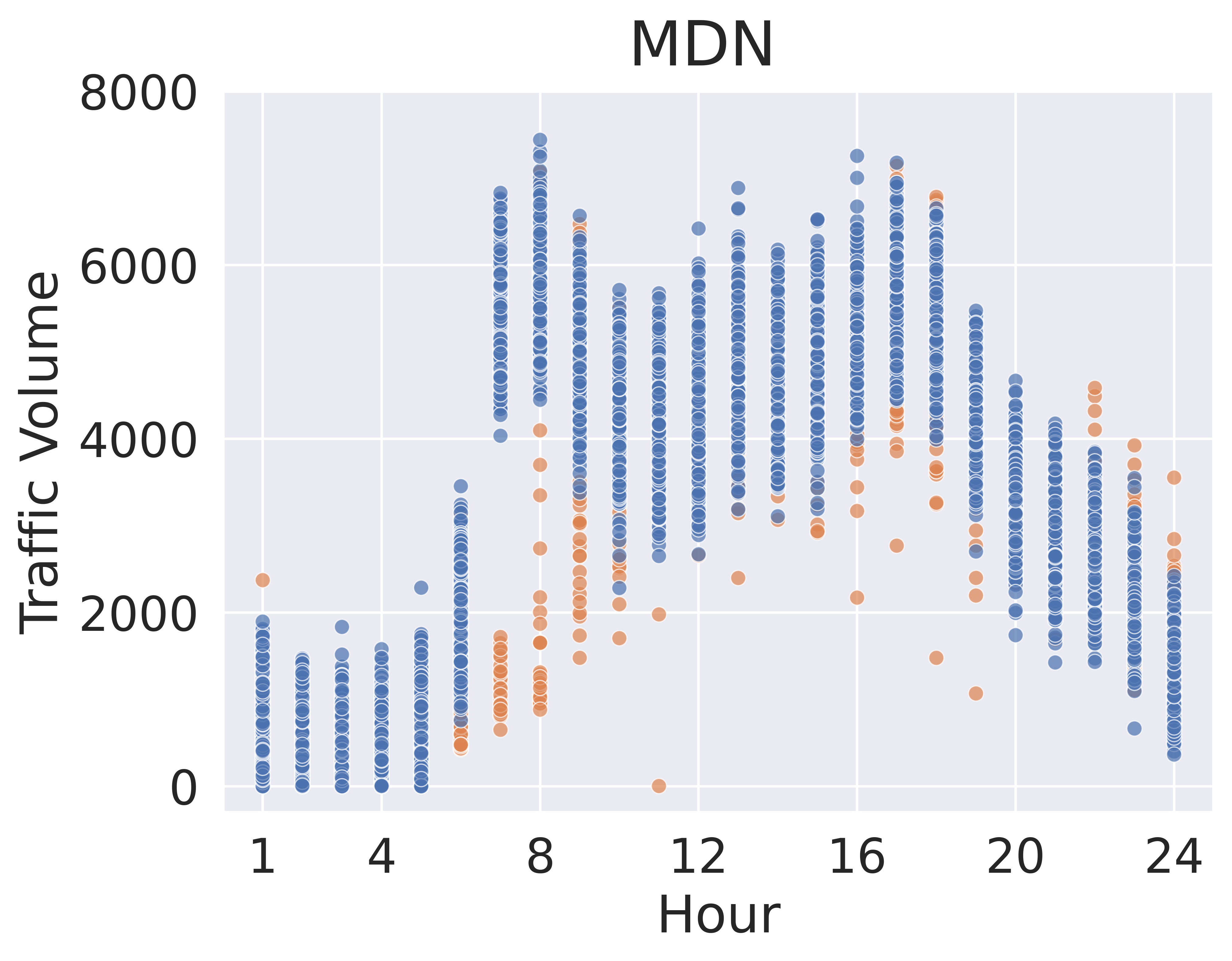}
     \end{subfigure}
          \begin{subfigure}[b]{0.245\textwidth}
         \includegraphics[trim={5 5 5 5}, clip,width=\textwidth]{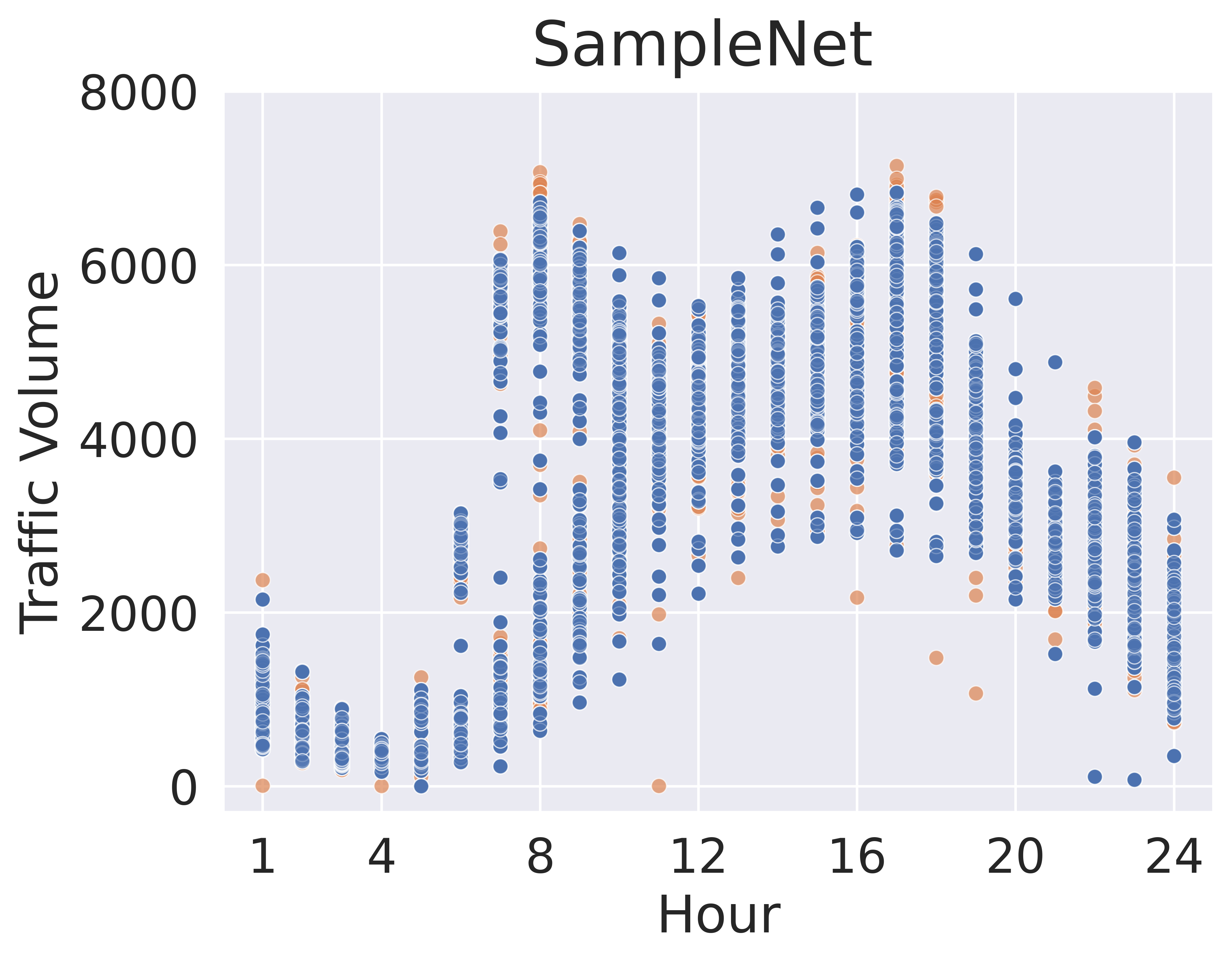}
     \end{subfigure}
     \caption{Scatter plots of samples (\textbf{blue}) from SampleNet and baselines on the Traffic dataset. \textbf{Test set} groundtruth samples (\textbf{orange}) are shown in the leftmost plot.}
     \label{fig:traffic_qualitative}
\end{figure*}

\textbf{Traffic Dataset}: We use the publicly available the Metro Interstate Traffic Volume Data Set from the UCI benchmark\footnote{\url{https://archive.ics.uci.edu/ml/datasets/Metro+Interstate+Traffic+Volume}}, which contains hourly traffic data from Minneapolis, Minnesota, USA between the years 2012 and 2018. We use hourly traffic volume data between the hours of 00:00 and 23:00 sampled from all days of the year (weekdays + weekends). Because of vast differences in hourly traffic volume on different days of the week, this dataset is naturally multimodal (\figref{fig:traffic_qualitative}). We train on data points corresponding to $4373$ hours from the year 2015, and test on all data points from the remaining $6$ years between the years 2012 and 2018.

On both datasets, we train all neural networks until for a maximum of $5000$ iterations for weather dataset and $30000$ iterations for the traffic dataset, or until convergence (SampleNet needs only $5000$ iterations on the traffic dataset to converge). We use a learning rate of $0.001$ for the Weather dataset and $0.01$ for the Traffic dataset. Since our goal behind these experiments is to show SampleNet's capacity to learn arbitrary distributions and not to compare performance with other baselines, we analyze performance and tune hyperparameters directly on the test set. All experiments are repeated over $5$ random seeds. Since the data is multimodal, we only compare SampleNet against MDN and $\beta$-NLL using the Energy Score, using the Gaussian NLL would not reflect the quality of predicted multimodal distributions. Results from the remaining baselines are all worse-performing, as can be seen when looking at the ES on the Weather dataset in Table~\ref{table:weather_proof_of_baselines}. For MDN, we use $J=5$. For $\beta$-NLL, we use $\beta=0.0$. For SampleNet, we use a Gaussian prior, and study the impact of the rest of the hyperparameters on performance in \secref{sec:experiments_results}. 

\begin{table}[t]
   \centering
   \caption{Results of more baselines on the Weather Dataset}
   \resizebox{0.45\textwidth}{!}{
  \begin{tabular}{cccc}
    \label{table:weather_proof_of_baselines}
    Dropout & $\beta$-NLL & Dropout \& $\beta$-NLL & Ensemble\\\midrule
    1.994 $\pm$0.423
 &0.594 $\pm$ 0.032
& 1.540 $\pm$ 0.828 
& 0.600 $\pm$ 0.011 \\\midrule \midrule
Student-t & MDN & SampleNet ($\eta=0$)& SampleNet ($\eta=5$) \\ \midrule
1.095 $\pm$ 0.057
& 0.210 $\pm$ 0.065
& 0.117 $\pm$ 0.002
&\textbf{0.106} $\pm$\textbf{0.002}
\\
\bottomrule
\end{tabular}}
\end{table}
\begin{table*}[t]
   \centering
   \caption{Regression dataset parameters.}
   \resizebox{0.7\textwidth}{!}{
  \begin{tabular}{lccc ccc}
    \label{table:regression_datasets_properties}
     Dataset& N & $D_{in}$ & $D_{out}$ & Num Neurons & Max Training Steps & $n_{\text{splits}}$\\
    \cmidrule(lr){1-4} \cmidrule(lr){5-7}
    Boston& 506 & 13 & 1 & 25 & 20000 & 20\\
    Kin8nm& 8192 & 8 & 1 & 50 & 100000 & 20\\
    Power& 9568 & 4 & 1 &  50 & 100000 & 20\\
    Yacht& 307 & 7 & 1 & 25 & 20000 & 20\\
    Concrete& 1030 & 8 & 1 & 25 & 20000 & 20 \\
    Wine (red)& 1599 & 11 & 1 & 25 & 20000 & 20 \\
    Wine (white)& 4898 & 11 & 1 & 25 & 20000 & 20\\
    Naval& 11934 & 16 & 1 &  50 & 100000 & 20\\
    Superconductivity& 21263 & 81 & 1&  50 & 100000 & 20\\
    Protein& 45730 & 9 & 1  & 100 & 100000& 5 \\
    Year& 515345 & 90 & 1 & 100 & 100000& 5 \\
    ObjectSlide& 136800 & 5 & 1  & 50 & 5000 & 20 \\
    Energy& 768 & 8 & 2 &  25 & 20000 & 20\\
    Carbon& 10721 & 5 & 3 &  50 & 100000 & 20\\    
    Fetch-Pick\&Place& 638400 & 29 & 3 & 50 & 5000 & 20\\
    \bottomrule
   \end{tabular}}
\end{table*}

\begin{table}[t]
   \centering
   \caption{The hyperparameter choices that produce the best results for SampleNet on real-world regression datasets.}
   \resizebox{0.45\textwidth}{!}{
  \begin{tabular}{lcccccc}
    \label{table:samplenet_regression_hyperparameters}
     Dataset& $M$ & $K$ & $L$ & $\eta$ & Prior\\\midrule
    Boston& 200&100 &4 &0.1 &Uniform\\
    Kin8nm &100 &100 &1 & 0.5& Uniform \\
    Power &50 &50 & 1& 0.5& Uniform\\
    Yacht &50 &50 & 1& 0.0& -\\
    Concrete& 50&50 &1 &0.0 &- \\
    Wine (red) &100 &100 &1 & 0.5&Uniform \\
    Wine (white) & 100& 50&1& 0.1& Uniform\\
    Naval& 200&50 &2 &0.0 &- \\
    Superconductivity &100 &100 & 1 &0.0 &- \\
    Protein  &100 & 100&1 &0.1&Gaussian  \\
    Year  &50 & 50&1 &1.0& Gaussian \\
    ObjectSlide  &200 &200 &1 &0.0&-  \\
    Energy &50 &50 &1 &0.1 & Uniform  \\
    Carbon &200 &100 &4 &0.1 & Uniform \\    
    Fetch-Pick\ &100&100 & 1& 0.1& Uniform\\
    \bottomrule
   \end{tabular}}
\end{table}

\begin{figure*}[t]
   \centering
         \begin{subfigure}[b]{0.32\textwidth}
    \includegraphics[width=\textwidth]{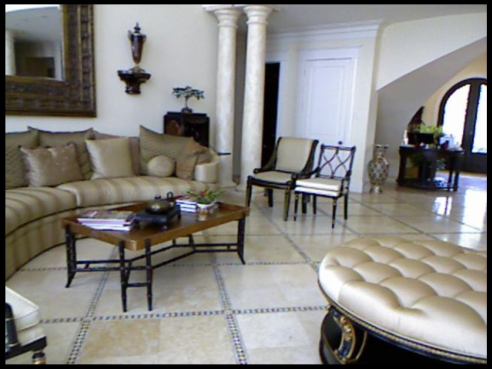}
     \end{subfigure}
    \begin{subfigure}[b]{0.32\textwidth}
    \includegraphics[width=\textwidth]{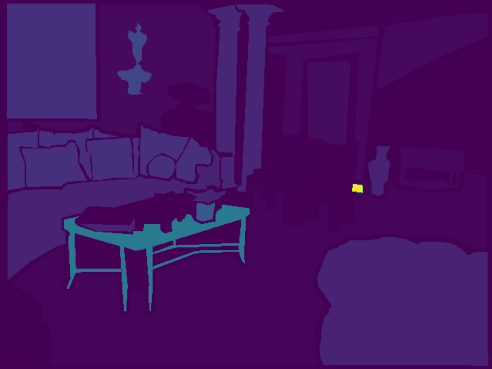}
     \end{subfigure}
     \begin{subfigure}[b]{0.32\textwidth}
    \includegraphics[width=\textwidth]{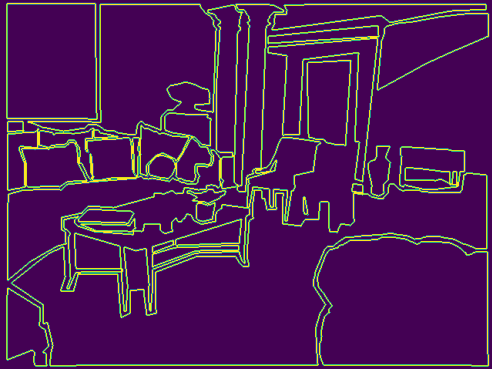}
     \end{subfigure}
    \caption{\textbf{Left}: A sample test image from the NYUv2 dataset~\cite{nyuv2}. \textbf{Middle}: Associated label mask. \textbf{Right}: Border pixels extracted using the label mask edges for evaluation.}
     \label{fig:depth_border_eval}
\end{figure*}
%%%%%%%%%%%%%%%%%%%%%%%%%%%%%%%
% UCI + 1dslide + fpp datasets%
%%%%%%%%%%%%%%%%%%%%%%%%%%%%%%%
\subsection{UCI Regression Datasets and Dynamics Models Datasets}
\label{appendix:regression_datasets}
Following the commonly used testing protocols in~\cite{skafte2019reliable,seitzer2022on}, we use 13 datasets from the UCI ~\cite{uci} benchmark. We also use 2 robot dynamics models datasets, ObjectSlide and Fetch-PickAndPlace, used to evaluate the performance of predictive distributions in~\cite{seitzer2022on}. Unlike~\cite{seitzer2022on}, only inputs and not outputs are whitened on the training set. We use a single-layer neural network with ELU activation. For all datasets, we randomly split the data into $n_{\text{splits}}$ train-test splits at $80\% - 20\%$ ratio. We further split each training set into a train/validation split and perform hyperparameter optimization on the validation set with early stopping using NLL for all baselines and the ES for SampleNet. We also determine the best learning rate for each configuration from the set \{1e-4, 3e-4, 7e-4, 1e-3, 3e-3, 7e-3\}. After determining the best configuration, we retrain the best model on the train set and generate the RMSE, NLL, and ES on the test set. \textit{We repeat this process for all splits and report the mean and standard deviation. We also perform performed a two-sided Kolmogorov-Smirnov test with a} $p\leq0.05$ \textit{significance level to determine the statistical significance of the results.}

Table~\ref{table:regression_datasets_properties} provides the number of data points $N$, input Dimensions $D_{in}$, output Dimensions $D_{out}$, number of neurons, the maximum number of training steps, and number of train-test splits $n_{\text{splits}}$ used for every real-world regression dataset used in our experiments. We train all neural networks using a minibatch size of 256 and a single Nvidia A100 GPU.

%%%%%%%%%%%%%%%%%%%%%%%%%%%%
% Depth Regression Datasets%
%%%%%%%%%%%%%%%%%%%%%%%%%%%%
\subsection{Monocular Depth Prediction Dataset}
\label{appendix:depth_prediction_datasets}
% Dataset description
The NYUv2~\cite{nyuv2} dataset consists of images and associated depth maps with a resolution of $640 \times 480$ and contains 120K training samples and 654 testing samples. We train using the 24K image-depth pairs subset provided by Lee~\etal\cite{lee2019big}. Following Seitzer~\etal\cite{seitzer2022on}, we use a simplified version of Adabins~\cite{bhat2021adabins}, removing the Mini-ViT Transformer module and only using the remaining encoder-decoder block. Each method replaces the last convolution of the encoder-decoder block with the appropriate output dimension for uncertainty prediction per pixel. The networks output a prediction with a resolution of $320 \times 240$ which is upsampled $2\times$ using nearest neighbors for SampleNet and bilinear interpolation for $\beta$-NLL, MDN, and Student-t. We train the networks using the ES for SampleNet and NLL for all other methods. SampleNet predicts 25 samples at each prediction pixel. We train all networks with a fixed seed and a batch size of 12 on 4  Nvidia A100 GPUs.

% Depth metrics.
We use the standard metrics originally proposed by Eigen~\etal~\cite{eigen2014depth}. Let $y_p$ be the ground truth depth for pixel $p$ and $\hat{y}_p$ be the predicted output of the model. The threshold accuracy ($\delta_i, i=1,2,3$) is the percentage of predicted pixels $\hat{y}_p$ s.t. $\max (\frac{y_p}{\hat{y}_p}, \frac{\hat{y}_p}{y_p}) < \delta_i$ where $\delta_1 = 1.25^1, \delta_2 = 1.25^2, \delta_3 = 1.25^3$. The average relative error (REL) is $\frac{1}{n}\sum^n_p{\frac{|y_p - \hat{y}_p|}{y_p}}$, the squared relative difference (Sq Rel) is $\frac{1}{n}\sum^n_p{\frac{\|y_p - \hat{y}_p\|^2}{y_p}}$, the root mean squared error (RMS) is $\sqrt{\frac{1}{n}\sum^n_p{(y_p - \hat{y}_p)^2}}$, the root mean squared log error (RMS log) is $\sqrt{\frac{1}{n}\sum^n_p{\|\log y_p - \log \hat{y}_p \|^2}}$, the average log10 error is $\frac{1}{n}\sum^n_p{|\log_{10} y_p - \log_{10} \hat{y}_p|}$, and the scale-invariant error (SI log) $\frac{1}{n^2} \sum_{i,j} \left(\left(\log \hat{y}_i - \log \hat{y}_j \right) - \left(\log y_i - \log y_j \right) \right)$. We additionally evaluate the methods on a border pixels subset. Along with depth maps, the NYUv2 dataset provides object label masks. As shown in \figref{fig:depth_border_eval}, we use the label masks to extract a subset of pixels along the border of objects to examine the behaviour of the methods on multiple depth modes. The NYUv2 dataset also provides instance maps, but both label and instance masks produce similar quantitative results. Based on visual inspection, label masks are chosen for better borders around objects.

\subsection{Compute Required to Generate Results}
\label{appendix:compute}
Our experiments and hyperparameter optimization were performed on an internal cluster with Nvidia A100 GPUs. Over the course of this work, we used 0.54 GPU years and 2.16 CPU core years. That equates to running a single A100 GPU and a 4 core CPU constantly for six and a half months.

%%%%%%%%%%%%%%%%%%%%%%%%%%%%%%%%%%%%%%%%%%%%%%%%%%%%%
% Any Additional Results that Don't fit in main paper%
%%%%%%%%%%%%%%%%%%%%%%%%%%%%%%%%%%%%%%%%%%%%%%%%%%%%%
\clearpage
\newpage

\subsection{Additional Results}
\label{appendix:additional_results}
\begin{figure}[H]
   \centering
         \begin{subfigure}[b]{0.23\textwidth}
    \includegraphics[width=\textwidth]{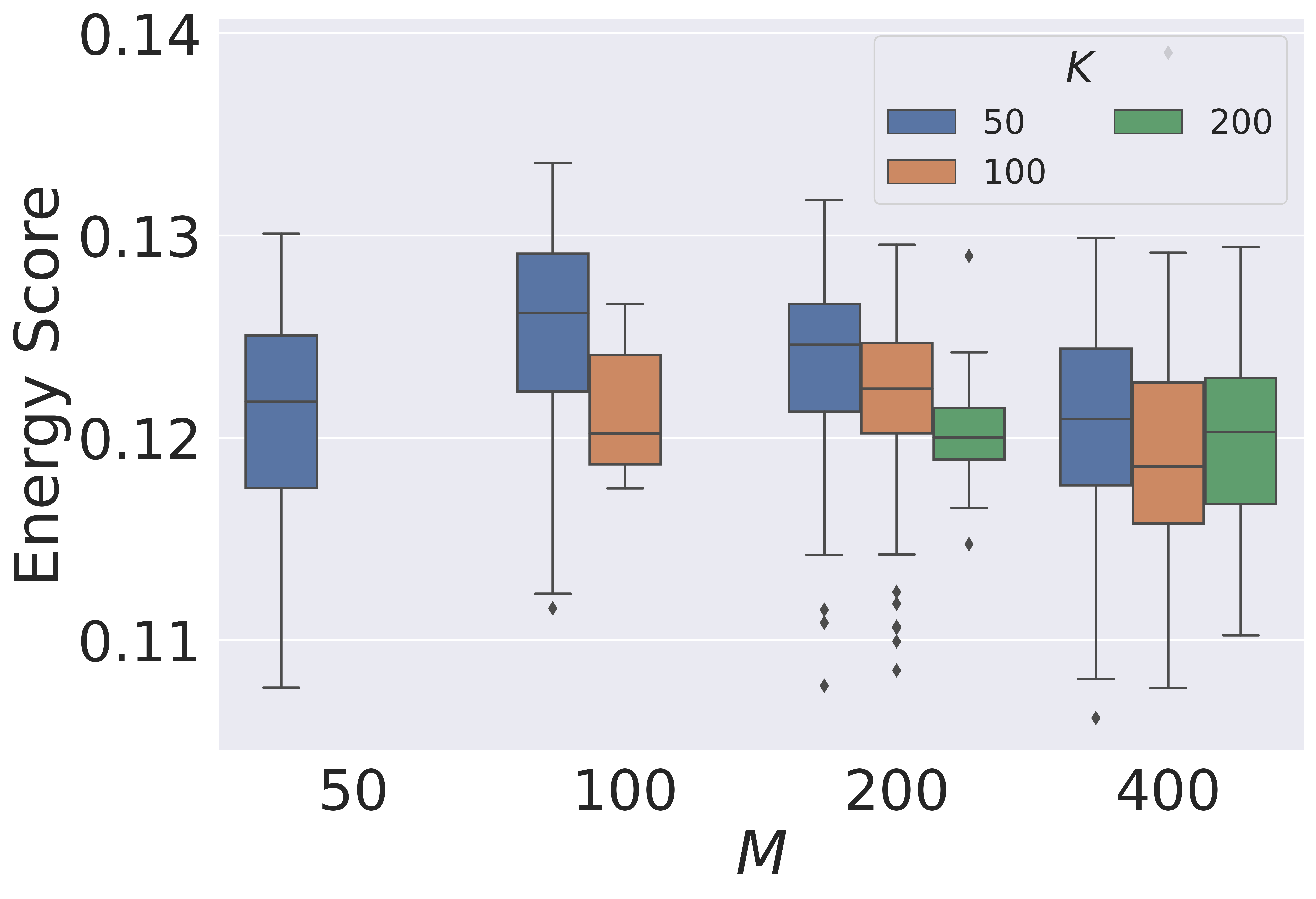}
     \end{subfigure}
    \begin{subfigure}[b]{0.23\textwidth}
    \includegraphics[width=\textwidth]{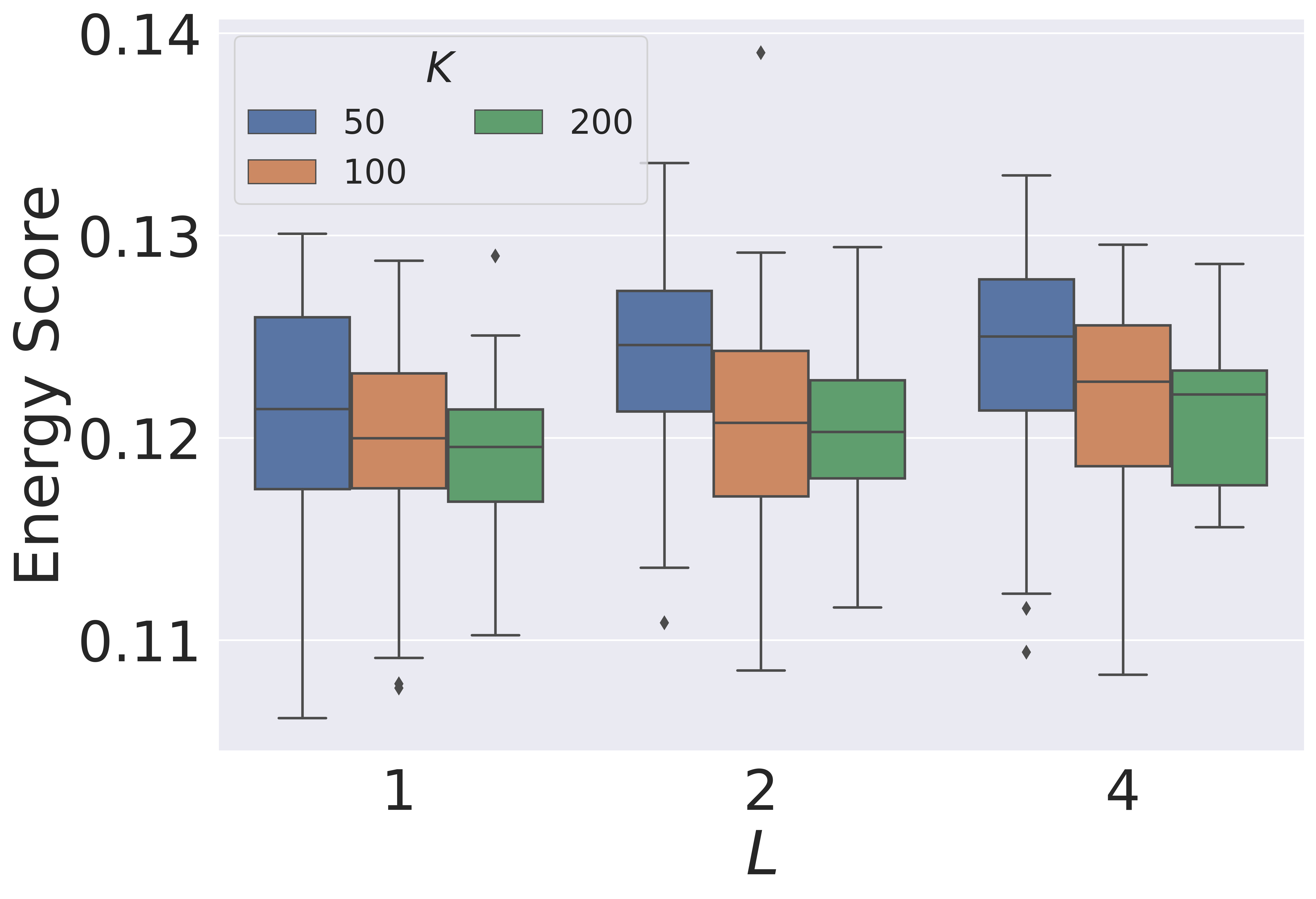}
     \end{subfigure}
    \caption{Additional hyperparameter results on the Weather Dataset. \textbf{Left}: box plot of the ES against $M$ for various values of minibatch loss sample size $K$. Results are aggregated over all values of $\eta$ and $L$. \textbf{Right}: box plot of the ES against $L$ for various values of the number of repetitions $K$. Results are aggregated over all values of $\eta$ and $M$.}
     \label{fig:ablations_weather_additional}
\end{figure}
\begin{figure}[H]
   \centering
    \includegraphics[width=0.45\textwidth]{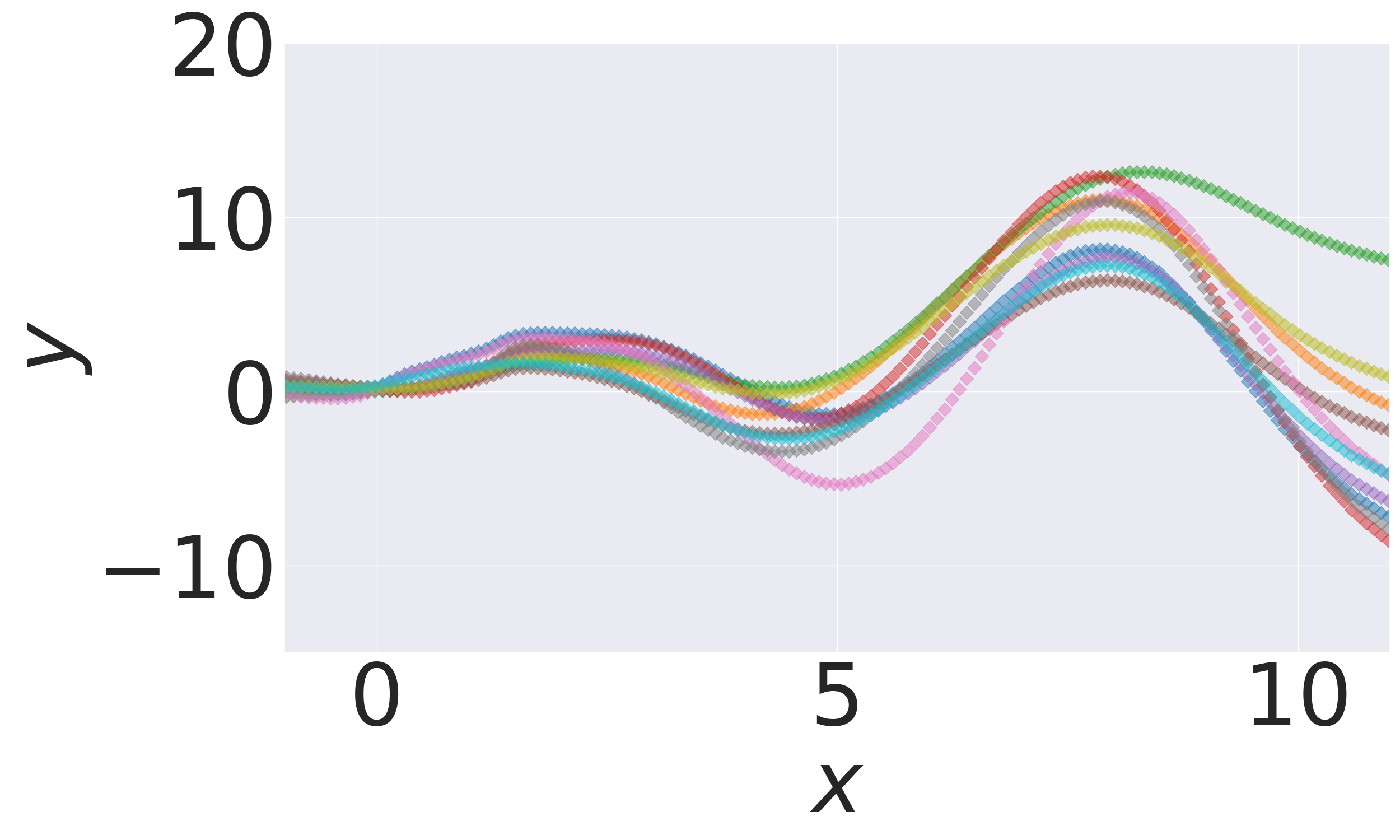}
    \caption{Plot of 10 different output neurons from SampleNet showing 10 learned functions on the toy example in~\secref{appendix:toy_data}}.
     \label{fig:function_colors}
\end{figure}
\begin{table*}[t]
   \centering
   \caption{RMSE results on real-world regression datasets. Best performing methods are marked in bold. The $\dagger$ symbol indicates a SampleNet trained without OT regularization ($\eta=0$).}
   \resizebox{\textwidth}{!}{
  \begin{tabular}{lccc cccccc}
    \label{table:rmse_real_regression}
     Dataset& N & $D_{in}$ & $D_{out}$ & Dropout & $\beta$-NLL & Dropout \& $\beta$-NLL & Student-t & MDN & SampleNet\\
    \cmidrule(lr){1-4} \cmidrule(lr){5-10}
    Boston& 506 & 13 & 1 & $\textbf{2.919}\pm \textbf{0.627}$& $3.707\pm 0.921$  &  $\textbf{3.056}\pm \textbf{0.732}$ & $\textbf{3.211} \pm \textbf{0.82}$ & $\textbf{3.168}\pm \textbf{0.698}$ & $\textbf{3.088} \pm\textbf{0.930}$\\
    Kin8nm& 8192 & 8 & 1 & $ \textbf{0.099}\pm \textbf{0.003}$& $\textbf{0.081}\pm  \textbf{0.003}$  &  $ \textbf{0.104}\pm\textbf{0.004 }$ & $ \textbf{0.087} \pm\textbf{0.003 }$ & $ \textbf{0.077} \pm \textbf{0.003}$ & $ \textbf{0.077}\pm \textbf{0.003}$\\
    Power& 9568 & 4 & 1 & $ 6.439\pm 0.260$& $ \textbf{4.148}\pm  \textbf{0.166}$  &  $ 8.527\pm 0.329$ & $ \textbf{4.157}\pm \textbf{0.143}$ & $\textbf{4.224} \pm \textbf{0.173}$ & $ \textbf{4.193}\pm \textbf{0.162}$\\
    Yacht& 307 & 7 & 1 & $ 1.473\pm  0.486$& $ 6.634\pm  1.123 $  &  $ 1.024\pm 0.39$ & $ 8.704 \pm 1.707$ & $ \textbf{1.772} \pm \textbf{0.582}$ & $ \textbf{0.446} \pm \textbf{0.189}^\dagger$\\
    Concrete& 1030 & 8 & 1 & $ \textbf{5.634}\pm \textbf{0.444}$& $ 6.232\pm 0.515 $  &  $ 6.354\pm 0.444$ & $ 6.106\pm0.495 $ & $ \textbf{5.897}\pm \textbf{0.691}$ & $ \textbf{5.526}\pm \textbf{0.590}^\dagger$\\
    Wine (red)& 1599 & 11 & 1 & $ \textbf{0.639}\pm \textbf{0.034}$& $\textbf{0.646}\pm \textbf{0.039} $  &  $ \textbf{0.642}\pm\textbf{0.033}  $ & $ \textbf{0.645}\pm \textbf{0.035}$ & $ \textbf{0.657}\pm \textbf{0.038}$ & $ \textbf{0.638}\pm \textbf{0.038}$\\
    Wine (white)& 4898 & 11 & 1 & $ \textbf{0.714}\pm \textbf{0.020}$& $ \textbf{0.715}\pm \textbf{0.018} $  &  $  \textbf{0.714}\pm\textbf{0.023} $ & $ \textbf{0.705}\pm \textbf{0.020}$ & $ \textbf{0.711}\pm \textbf{0.021}$ & $ \textbf{0.701}\pm \textbf{0.018} $\\
    Naval& 11934 & 16 & 1 & $0.012\pm 0.001$& $0.006\pm 0.007 $  &  $  0.015\pm 0.001 $ & $ -7.873\pm0.659  $ & $ 0.006\pm 0.008$ & $ \textbf{0.001} \pm \textbf{1e-5}^\dagger$\\
    Superconductivity& 21263 & 81 & 1& $ \textbf{12.82}\pm \textbf{0.327}$& $ 14.240\pm 0.361 $  &  $ 14.266\pm 0.313$ & $ 14.019\pm 0.318$ & $ 14.337\pm0.332 $ & $ 13.719\pm 0.288^\dagger$\\
    Protein& 45730 & 9 & 1 & $ 4.508\pm0.047 $& $ \textbf{4.413}\pm \textbf{0.020} $  &  $ 4.696\pm 0.021$ & $ 4.524\pm 0.035$ & $ \textbf{4.384}\pm \textbf{0.042}$ & $ \textbf{4.434}\pm \textbf{0.030}$\\
    Year& 515345 & 90 & 1 & $23.192\pm 0.870$& $ \textbf{9.690}\pm  \textbf{0.513}$  &  $  30.263\pm 0.677$ & $ \textbf{9.423}\pm\textbf{0.149} $ & $ \textbf{10.619}\pm \textbf{1.181}$ & $ \textbf{9.856}\pm \textbf{0.306}$\\
    ObjectSlide& 136800 & 5 & 1 & $ 0.0305\pm 0.001 $& $ 0.033\pm 0.001 $  &  $0.027 \pm0.002 $ & $ 0.025\pm 0.001$ & $ 0.0243\pm 0.001$ & $ 0.0313\pm0.002^\dagger $\\
    Energy& 768 & 8 & 2 & $ 1.401\pm 0.159$& $ 1.363\pm  0.232$  &  $ 3.124\pm 0.223 $ & $ 1.382\pm0.324 $ & $ 1.942\pm0.285 $ & $\textbf{1.144}\pm \textbf{0.111}$\\
    Carbon& 10721 & 5 & 3 & $ 0.011\pm 0.002$& \textbf{85e-4}$  \pm $ \textbf{22e-4} &  $ 0.013\pm  0.002$ & \textbf{89e-4}$ \pm $\textbf{21e-4}& \textbf{82e-4}$ \pm $ \textbf{22e-4}& \textbf{85e-4}$\pm $\textbf{22e-4}\\    
    Fetch-Pick\&Place& 638400 & 29 & 3 &37e-4 $ \pm $12e-5& 31e-4$ \pm  $5e-5  &  33e-4$ \pm $8e-5 & 31e-4$ \pm $ 6e-5& 38e-4$\pm $3e-5 & \textbf{30e-4}$\pm $\textbf{2e-5}\\
    \bottomrule
   \end{tabular}}
\end{table*}

\begin{table*}[t]
\centering
\caption{Additional results on the NYUv2 dataset with varying number of output samples $M$. $K$ is fixed to be 25 and $L$ = 1. It can be noted that the performance of SampleNet is stable as the number of samples increases.}
\label{table:depth_prediction_results_samples}
\resizebox{\textwidth}{!}{
\begin{tabular}{lcc ccccccccc cc cc}
& \multicolumn{10}{c}{Deterministic Metrics} & \multicolumn{2}{c}{Scoring Rules}& \multicolumn{2}{c}{Average Var} \\
 Method & $\eta$ & $M$ & $\delta$1↑   & $\delta$2↑  & $\delta$3↑  & REL↓     & Sq Rel↓   & RMS↓     & RMS log↓   & log10↓  & SI log↓          & NLL↓            & ES↓ &  All & Border\\ \cmidrule(lr){1-3}\cmidrule(lr){4-12}\cmidrule(lr){13-14}\cmidrule(lr){15-16}
 Varnet-ES & N/A  &25& 0.8909    & 0.9813          & 0.9957          & 0.1065          & 0.0618          & 0.3755          & 0.1372          & 0.0452          & 11.306          & 18.593 & 0.2341  & 0.0123& 0.0258 \\\cmidrule(lr){1-3}\cmidrule(lr){4-12}\cmidrule(lr){13-14}\cmidrule(lr){15-16}
SampleNet & 0 & 25    & 0.8891          & 0.9800          & 0.9950          & 0.1071          & 0.0642          & 0.3795          & 0.1383          & 0.0456          & 11.332          & 1.397 & 0.2371  & 0.0157 & 0.0340 \\
SampleNet & 0 & 50    & 0.8909          & 0.9799          & 0.9955          & 0.1070          & 0.0645          & 0.3790          & 0.1375          & 0.0454          & 11.265          & 1.399 & 0.2381 & 0.0132 & 0.0291 \\
SampleNet & 0 & 100    & 0.8868          & 0.9795          & 0.9952          & 0.1077          & 0.0637          & 0.3875          & 0.1396          & 0.0462          & 11.449          & 1.394 & 0.2392  & 0.1332 & 0.1735 \\
SampleNet & 0.01 & 25    & 0.8889          & 0.9802          & 0.9953          & 0.1085          & 0.0649          & 0.3831          & 0.1394          & 0.0459          & 11.460          & 1.398 & 0.2407  & 0.0115 & 0.0261 \\
SampleNet & 0.1 & 25    & 0.8846          & 0.9792          & 0.9949          & 0.1096          & 0.0642          & 0.3840          & 0.1398          & 0.0464          & 11.316          & 1.401 & 0.2429  & 0.0124 & 0.0278 \\
SampleNet & 0.5 & 25    & 0.8850          & 0.9794          & 0.9956          & 0.1090          & 0.0644         & 0.3869          & 0.1401          & 0.0464          & 11.383          & 1.399 & 0.2438  & 0.0124 & 0.0286 \\

\bottomrule
\end{tabular}}
\end{table*}

\begin{figure*}[t]
  \centering
    \begin{subfigure}[b]{0.4\textwidth}
    \includegraphics[width=\textwidth]{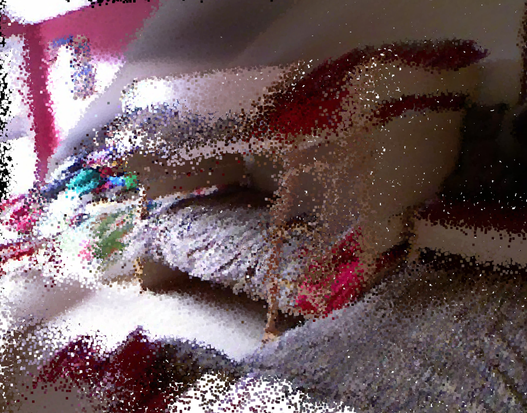}
     \end{subfigure}
    \begin{subfigure}[b]{0.4\textwidth}
         \includegraphics[width=\textwidth]{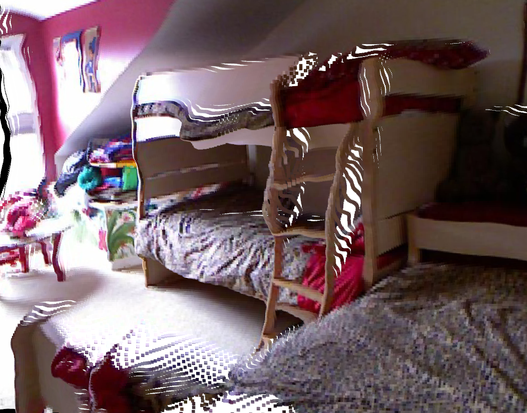}
     \end{subfigure}
    \begin{subfigure}[b]{0.4\textwidth}
    \includegraphics[width=\textwidth]{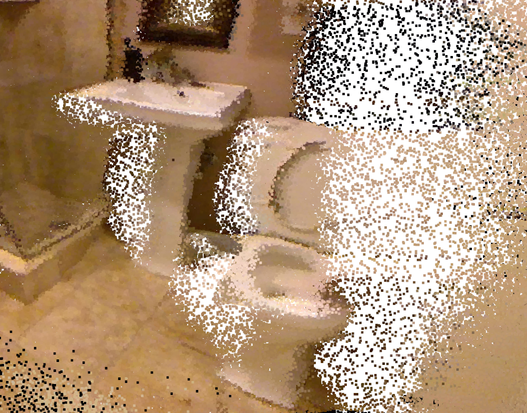}
     \end{subfigure}
    \begin{subfigure}[b]{0.4\textwidth}
         \includegraphics[width=\textwidth]{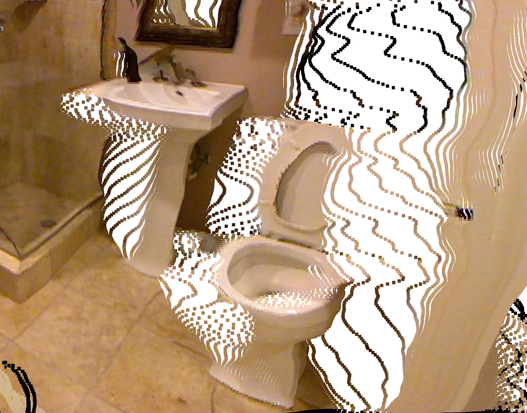}
     \end{subfigure}
    \begin{subfigure}[b]{0.4\textwidth}
    \includegraphics[width=\textwidth]{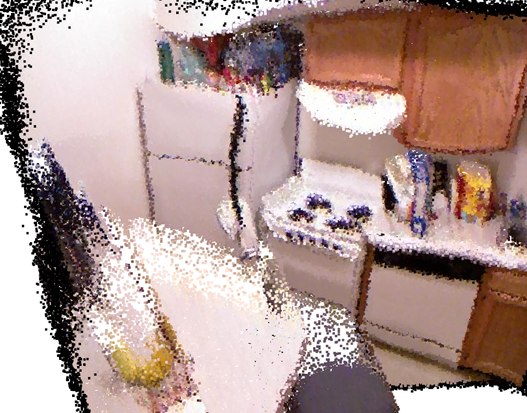}
     \end{subfigure}
    \begin{subfigure}[b]{0.4\textwidth}
         \includegraphics[width=\textwidth]{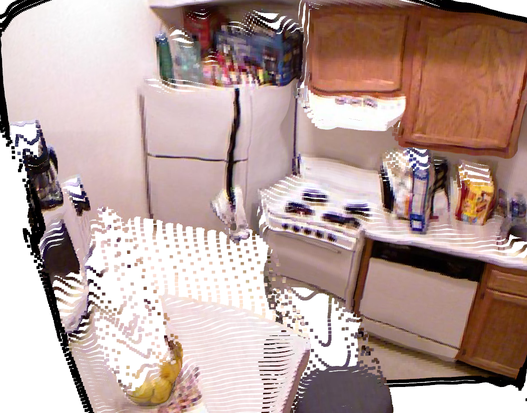}
     \end{subfigure}
    \begin{subfigure}[b]{0.4\textwidth}
    \includegraphics[width=\textwidth]{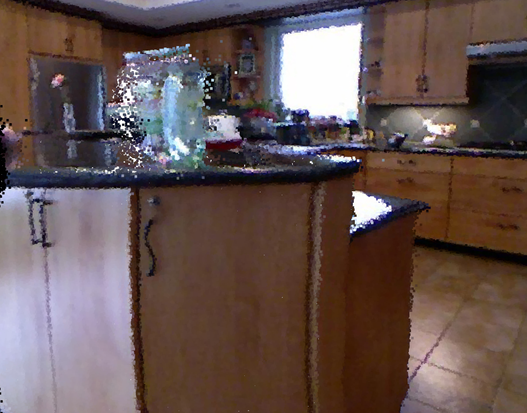}
     \end{subfigure}
    \begin{subfigure}[b]{0.4\textwidth}
         \includegraphics[width=\textwidth]{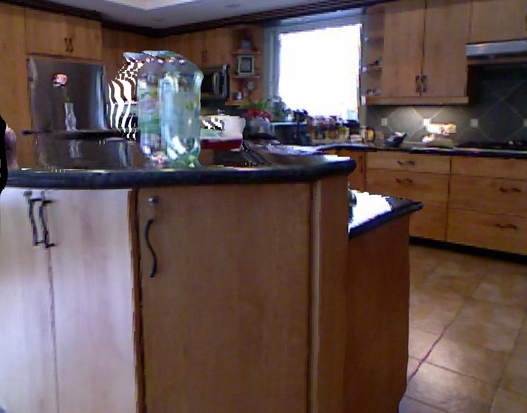}
     \end{subfigure}
    \caption{Qualitative monocular depth prediction results as point clouds from the NYUv2 dataset~\cite{nyuv2}. \textbf{Left}: $\beta$-NLL with $\beta = 1$. \textbf{Right}: SampleNet provides sharper distributions, resulting in a more geometrically consistent point cloud compared to other methods.}
    \label{fig:depth_pc}
\end{figure*}

\end{document}